\documentclass{article}
\pdfoutput=1

\usepackage{arxiv}

\usepackage[utf8]{inputenc} % allow utf-8 input
\usepackage[T1]{fontenc}    % use 8-bit T1 fonts
\usepackage{hyperref}       % hyperlinks
\usepackage{url}            % simple URL typesetting
\usepackage{booktabs}       % professional-quality tables
\usepackage{amsfonts}       % blackboard math symbols
\usepackage{nicefrac}       % compact symbols for 1/2, etc.
\usepackage{microtype}      % microtypography
\usepackage{lipsum}		% Can be removed after putting your text content
\usepackage{graphicx}
\usepackage{xltabular}
\usepackage{natbib}
\usepackage{doi}
\usepackage{xspace,mfirstuc,tabulary}
\usepackage{graphicx}
\usepackage{textgreek}
 \usepackage{rotating}
        \usepackage{supertabular} 
        \usepackage{multirow}
        \usepackage{array}
\usepackage{rotating}
\usepackage{makecell, multirow}
\usepackage[T2A]{fontenc}
\usepackage[utf8]{inputenc}
\usepackage[english]{babel}
\newcommand{\spedac}{\textsc{SPeDaC}}

\title{Is Your Model Sensitive? SPeDaC: A New Resource and Benchmark for Training Sensitive Personal Data Classifiers}

\author{ \href{https://orcid.org/0000-0002-7697-2848}{\includegraphics[scale=0.06]{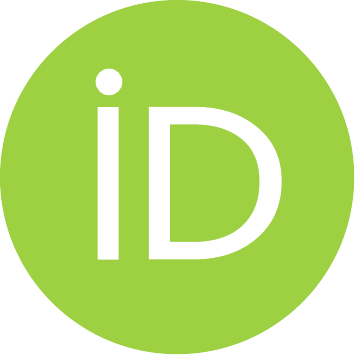}\hspace{1mm}Gaia~Gambarelli}\\
	Department of FICLIT \\
	Ellysse srl. \\
	University of Bologna, Italy \\
	Reggio Emilia, Italy \\
	\texttt{gaia.gambarelli2@unibo.it} \\
	\And
	\href{https://orcid.org/0000-0001-5568-2684}{\includegraphics[scale=0.06]{orcid.pdf}\hspace{1mm}Aldo~Gangemi} \\
	Department of FICLIT \\
	IST-CNR \\
	University of Bologna, Italy \\
	Roma, Italy \\
	\texttt{aldo.gangemi@unibo.it} \\
		\And
	\href{https://orcid.org/0000-0003-2483-8564}{\includegraphics[scale=0.06]{orcid.pdf}\hspace{1mm}Rocco~Tripodi} \\
	Department of LILEC\\
	University of Bologna, Italy\\
	\texttt{rocco.tripodi@unibo.it} \\
}

\renewcommand{\shorttitle}{\textit{Is Your Model Sensitive?}}

\hypersetup{
pdftitle={Is Your Model Sensitive? SPeDaC: A New Resource and Benchmark for Training Sensitive Personal Data Classifiers},
pdfauthor={Gaia~Gambarelli, Aldo~Gangemi, Rocco~Tripodi},
pdfkeywords={Sensitive Personal Data, Transformer Models, Privacy Protection, RoBERTa, Classification task, Sensitive Data Corpus},
}

\begin{document}

\clearpage
\null
\thispagestyle{empty}
\begin{figure}
    \centering
    \includegraphics[width=17cm]{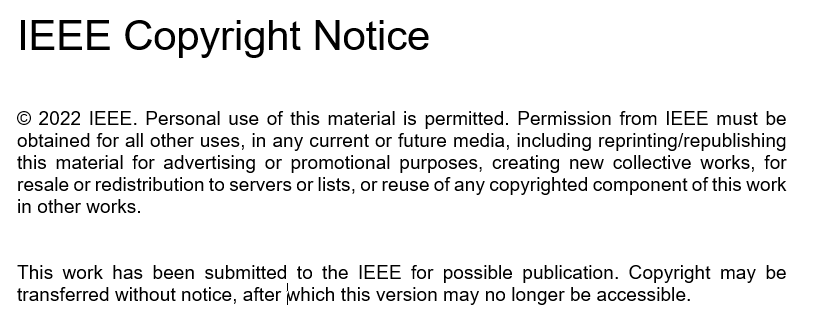}
\end{figure}

\clearpage

\maketitle

\begin{abstract}
In recent years, there has been an exponential growth of applications, including dialogue systems, that handle sensitive personal information. This has brought to light the extremely important issue of personal data protection in virtual environments. Sensitive Information Detection (SID) approaches different domains and languages in literature. However, if we refer to the personal data domain, a shared benchmark or the absence of an available labeled resource makes comparison with the state-of-the-art difficult. We introduce and release \spedac , a new annotated resource for the identification of sensitive personal data categories in the English language. \spedac enables the evaluation of computational models for three different SID subtasks with increasing levels of complexity. \spedac 1 regards binary classification, a model has to detect if a sentence contains sensitive information or not; whereas, in \spedac 2 we collected labeled sentences using 5 categories that relate to macro-domains of personal information; in \spedac 3, the labeling is fine-grained (61 personal data categories). We conduct an extensive evaluation of the resource using different state-of-the-art-classifiers. The results show that \spedac is challenging, particularly with regard to fine-grained classification. The transformer models achieve the best results (acc. RoBERTa on \spedac 1 = 98.20\%, DeBERTa on \spedac 2 = 95.81\% and \spedac 3 = 77.63\%).
%based on RoBERTa, a neural architecture that achieves strong performances on the detection of sensitive sentences (98.20\% acc.) and on personal data categories classification (94.24\% acc.), while it offers a benchmark result concerning the third dataset.
\end{abstract}


\section{Introduction}
\label{Introduction}
In recent years, there has been an exponential growth of applications, including dialogue systems that handle sensitive personal information \cite{Larson2021NotDS, Adhikari2018, Hendrickx2021}.
Identifiable individuals can explicitly or implicitly reveal inferable personal information from the texts they write and from the information they share daily online (in blogs, public pages, social media etc.). The context in which personal information can be expressed concerns not only public online environments but also private interactions, in which, sometimes, the sharing of such information is deemed necessary. Exchanges of emails in company structures, virtual interactions between users and operators of a Customer Service, or even the use of applications based on human-robot (H-R) interactions are all scenarios in which the management of personal information is important. 
In online conversations and unstructured text, for example, the loss of privacy can be very high and the average cost of data breaches has increases over the years \cite{IBMBreachReport}. The
loss of personal information to 3rd parties can have both legal and economic repercussions
on the users and managers of the service, and, in social terms, on the individuals directly
involved. Finally, it is estimated that 80\% of the data currently disseminated on the Internet
is of an unstructured type \cite{Allahyari2017}, data not present in a relational database, which can
be presented in an irregular and contextual form.

According to General Data Protection Regulation (GDPR, 2018) \cite{GDPROnline}, the right to privacy regarding sensitive personal data is claimed; managing privacy and understanding the processing of personal data has become a fundamental right, especially within the European Union (GDPR, Recital 6) \cite{GDPROnlineRecital}. 
Following the regulatory definition, `personal data' means `any information relating to an identified or identifiable natural person (`data subject'); an identifiable natural person is one who can be identified, directly or indirectly, in particular by reference to an identifier such as a name, an identification number, location data, an online identifier or to one or more factors specific to the physical, physiological, genetic,
mental, economic, cultural or social identity of that natural person' (GDPR, 4.1) \cite{GDPROnlineArt4}.

Consequently, many studies have focused on protecting privacy in virtual spaces from several points of view e.g., data sanitization and anonymization methods.
Models of data sanitization by using deletion operations on transactions are one of the most common approaches in  Privacy Preserving Data Mining (PPDM) \cite{Agrawal2000}. PPDM techniques `allow the extraction of information from data sets while preventing the disclosure of data subjects' identities or sensitive information' \cite{Esma2021}. Data sanitization generally aims to hide sensitive information applying minimal side effects and keeping the original database as authentic as possible \cite{Cheng2015}. Several methods are applied to input data e.g., data perturbation \cite{Kargupta2003}, \cite{Xiao2021}, cryptography \cite{Lu2014}, \cite{Yang2019}, and anonymity with different techniques methods \cite{Sweeney2002}, \cite{Machanavajjhala2006}, \cite{Li2007}, \cite{Abadi2016}.
A recent PPDM study \cite{Lin2021} introduces PACO2DT, an ACO-based multiobjective model which uses transaction deletion to hide sensitive and confidential information. The information to hide is defined by an expert in the industry in form of an input sensitive itemset, deleted and distributed to the IOT devices for configuration. In a previous study \cite{Lin2014}, where hiding-missing-artificial utility (HMAU) algorithm is introduced to address the PPDM problem, the authors propose in future work to extend the sensitive itemsets to be hidden to the sensitive association rules and to decrease the confidence of sensitive association rules. A type of extension is proposed introducing High-Utility Itemset Mining (HUIM), a model which discovers itemsets reporting a high profit in transaction databases \cite{Lin2016}. The authors introduce, as an extension of PPDM, the Preserving Utility Mining (PPUM) that hides Sensitive High-Utility Itemsets (SHUIs) considering their profit. PPDM and PPUM algorithms are included in the proposed interface Privacy-Preserving and Security Mining Framework (PPSF) \cite{Lin2018}.
One of the greatest risks of failure of these models is the loss of information. Even before the failure of the anonymization algorithm, there may be a missing identification of sensitive information.

Sensitive Information
Detection (SID) is a subpart of Data
Leak Detection (DLD) that deals with the automatic identification of sensitive information. The work also contributes to improving Data Loss Prevention (DLP) systems and industrial problems designed to help businesses to avoid data breaches, presenting a way to train, classify and perform the classification of sensitive text \cite{Hart2011}. Most of the current tools offer DLP services for the automatic identification of personally
identifiable information (PII) \cite{Google2022}, \cite{IBM2022}, \cite{Microsoft2022}. This paper addresses the challenge of identifying complex personal information in unstructured text. Words are sensitive or not sensitive depending on their context. Using different expressions in natural language, the same keyword can acquire a sensitive or non-sensitive character.

Related work in SID is often conducted in different domains or languages; however, frequently, there is no common benchmark or available labeled resources to compare the results with the state-of-the-art methods. We have attempted to fill this gap by introducing and evaluating a new sensitive resource. The datasets are freely available and can be reuse for training new models or as a benchmark to compare the results to state-of-the-art models.

At the same time, evaluating the resource, we have a neural networks method based on the transformer architecture \cite{Vaswani2017}, which has recently been used in SID tasks and has achieved astonishing performances on standard Natural Language Processing (NLP) tasks. The contributions of this study are as follows:

\begin{enumerate}
    \item we present \spedac (\underline{S}ensitive \underline{Pe}rsonal \underline{Da}ta \underline{C}ategories corpus): a benchmark built and manually labeled for personal data categories (PDCs). The dataset contributes to the detection of sensitive sentences and their classification as PDCs. We implicitly contribute to the evaluation gap \cite{Neerbek2020} and the absence of an available labeled resource concerning personal information;
    \item we report the results of several experiments conducted using different state-of-the-art models, including a classifier based on the transformer architectures \cite{Liu2019}. We aim to evaluate the \spedac dataset, propose a benchmark and analyze the validity of modern neural network approaches to the task of automatic identification of sensitive content. 
\end{enumerate}

The article consists of the following sections: Section \ref{Related work} is devoted to related work in the automatic identification of sensitive content and the use of transformer neural networks in text classification. In Section \ref{Materials and Models} we describe the materials and models involved: in Section \ref{Data Privacy Vocabulary (DPV)} the taxonomy we use to define the PDCs; in Section \ref{Dataset: a sensitive data corpus} \spedac, the constructed and labeled sensitive data corpus, is presented. The resource is evaluated in Section \ref{Models}, where we further explore the machine learning models as well as the transformer networks, both used to conduct comparative experiments and to validate the efficiency of the latter. In Section \ref{Experimental process}, we describe the experimental process which includes the feature extraction and the models set up, while in section \ref{Results} we report the results. Finally, Section \ref{Conclusion and Future Work} presents conclusions and future directions of work, and in Section \ref{Ethical Disclosure} you can find an ethical disclosure that concerns the protection from improper use of the resources presented.

\section{Related Work}
\label{Related work}
The domain of our study concerns personal data categories. In literature, not so many works focus on this specific domain, e.g., basic personal information \cite{Dias2020,Guo2021}, personal health information (PHI) \cite{Garcia2020}, ethnic origin and political opinion information \cite{Genetu2021}.

However, regardless of the type of information considered in the literature, sensitive data can be identified in a rigid and context-less manner or can be disambiguated or inferred from the context. We can divide the studies into two macro approaches: 
\begin{enumerate}
    \item \textbf{non-context-aware approach}, where sensitive information does not depend on the context in which it appears; for example, a word can be identified as sensitive regardless of the sentence in which it is used;
    \item \textbf{context-aware approach}, where the sensitivity of the data varies according to the context. Only given the sentence, we can infer the sensitivity of a given word.  In fact, assuming the textual unit in which a word appears as context  \cite{Neerbekthesis}, we consider the context of a word to be the sentence in which it appears. Nevertheless, this sensitive context can be extended to paragraphs or documents.
\end{enumerate}

The first non-context-aware approach includes work based on the identification of fixed context with n-gram techniques \cite{Hart2011,Islam2014} or rule-based inferences to identify contextless words with sensitivity scores, \cite{Chow2008,Geng2011}. The contextualized approach appears in literature with an embedding technique for the recognition of a fixed context \cite{Mcdonald2017}. 

Among the most recent works, we see the use of neural networks, for example recursive neural networks for automatic paraphrasing applied to the identification of sensitive data \cite{Neerbek2020}. Convolutional neural networks (CNNs) \cite{LeCun1989} have also been used for the sensitive detection of military and political documents in the Chinese language \cite{Xu2019}. Bidirectional Long-Short Term Memory (Bi-LSTM) neural networks \cite{Schuster1997} have been used in a study conducted in the Chinese language on unstructured text \cite{Lin2020} and for the identification of personal data in an Amharic text \cite{Genetu2021}. 

Finally, the field of identification of sensitive data began to take advantage of the transformer architecture \cite{Vaswani2017}. A study conducted in Spanish \cite{Garcia2020} used a BERT-based sequence labeling model to detect and anonymize sensitive data in the clinical domain. Specifically, they used two datasets of medical reports and ran comparison experiments using Conditional Random Field (CRF) and BERT. In the first dataset, the pre-trained BERT model outperformed the other systems, whereas, in the second, it fell at 0.3 F1-score points behind the shared task winning system, but the authors did not try more sophisticated fine-tuning strategies. A recent study on the English language \cite{Guo2021} proposed ExSense, a model named BERT-BiLSTM-attention for extracting sensitive information from unstructured text. The experimental process was conducted on the Pastebin dataset \cite{PastebinOnline}, manually labeled with personal information. Personal information refers to identifiable persons, such as name, address, date of birth, social security numbers (SSN), and telephone numbers. This model had an F1 score of 99.15\%. As the authors stated, ExSense can identify limited types of sensitive information.
The identified categories are, therefore, attributable to very specific entities, often presenting a fixed structure.

A novel framework, Just Fine-tune Twice (JFT), recently proposed \cite{Shi2022}, aims first to redact in-domain data of the sensitive task and fine-tune the model; second, to privately fine-tune the model on the original sensitive data. The first step allows the model to directly learn information from
the in-domain data and to work with a limited amount of data. The goal of the paper is to show the potential outcomes of JFT, and basic sensitive information is treated.

Recent literature on this task highlights the great potential of transformer-based models. However, the type of personal data investigated is often not very challenging. Transformer-based models have never been tested on the English language on such a broad domain of PDCs, such as the one presented in this work. Considering the definition of `personal data' given by the GDPR (Section \ref{Introduction}), many types of data can be identified textually. Categories such as names, addresses, and telephone numbers can be identified directly, through entities, while there are personal categories, such as health status, preferences, and social status, that can be more complex to identify or infer. Their common feature is that they can be directly or indirectly related to an identifiable person. In addition to investigating the accuracy of PDCs identification, we measured the accuracy that a transformer network can achieve in discriminating between sentences with and without sensitive content, based on the same potentially sensitive linguistic patterns occurring in different sentences that confer sensitivity or not.

Nevertheless, how can we identify the types of sensitive data categories to consider? The World Wide Web Consortium (W3C) \cite{W3COnline} created the Data Privacy Vocabulary (DPV) in 2019 \cite{Pandit2019}, a resource aimed at ensuring the interoperability of data privacy, which therefore represents a highly valid reference taxonomy \cite{DPVOnline}. We have used this as an authoritative reference to identify the categories of personal data (PDCs) to be analyzed. An extension of DPV regarding  extended personal data concepts was recently released \cite{DPV-PDOnline}. The resource will be discussed in more detail in Section \ref{Data Privacy Vocabulary (DPV)}. Regarding the second problem, our approach aims to be context-aware; the analysis is therefore a level-sentence, as we will describe in Section \ref{Dataset: a sensitive data corpus}.

As mentioned above, one of the major obstacles is the corpora and resources currently available to form and compare sensitive detection models \cite{Neerbek2020} . Some public corpora, that contain sensitive data and which have been used in the sensitive detection literature, are as follows: 
\begin{enumerate}
    \item the Enron email dataset \cite{EnronOnline}, which collects more than 600,000 e-mails from the American Enron Corporation, with approximately 2,720 documents manually labeled by human annotators, lawyers, and professionals in 2010. However, annotations only cover specific topics, such as business transactions, forecasts and projects, actions, and intentions. This dataset was used as an evaluation dataset in related studies \cite{Chow2008,Hart2011,Neerbek2020}.
    \item The Monsanto dataset \cite{MonsantoOnline}, published in 2017, consists of secret legal acts. This resource was similarly used for evaluation \cite{Neerbek2020}. 
    \item PII dataset from Pastebin \cite{Guo2021}. The authors collected documents from Pastebin, obtaining 144,967 text sequences as training data and identifying 4 types of PII information in text using regular expressions for content-based sensitive information and a BERT-BiLSTM attention model to automatically extract context-based sensitive information from the preprocessed text.
    \item Wikipedia dataset. Wikipedia articles or pages are very easy to acquire and contain different types of sensitive information. In a privacy ensuring study \cite{Hart2011} the authors have created a Wikipedia Test corpus, randomly sampling 10K Wikipedia articles. In another related work \cite{Sanchez2014a} the aim was to establish a framework for measuring the disclosure risk caused by semantically related terms; the authors used Wikipedia pages of individuals e.g., movie stars. They used a manual annotation for sentences on Wikipedia pages relating to PII  typically defined by keywords e.g., HIV (state of health), Catholicism (religion), and Homosexuality (sexual orientation). 
\end{enumerate}
The Enron corpus could be representative of organizational email conversations, including informal mails between colleagues. However, since it dates back to 2002, it cannot be considered very representative of today's communication style. Although more recent, the Monsanto dataset, is a domain-specific corpus that would barely cover many PDCs, other than those closely related to the legal domain. For these reasons, they could not represent a point of reference for the specific identification of personal data. The dataset from Pastebin is not currently available; furthermore, the investigated categories refer to PII, frequently detected through regular expressions or very narrow linguistic patterns. Even the Wikipedia dataset is not publicly available and, in any case, complex sensitive categories are not considered.

This brief survey highlights the clear lack of an available released labeled resource for the task of automatic identification of sensitive personal data. For this reason, this study aims to offer an evaluation and reusable resource as contribution.

\section{Materials and Models}
\label{Materials and Models}
In this section, we deepen the taxonomy used as a reference for the identification of the PDCs analyzed (Section \ref{Data Privacy Vocabulary (DPV)}), describe \spedac, the corpus built and evaluated (Section \ref{Dataset: a sensitive data corpus}) and introduce the machine learning and transformer network models used to conduct the classification experiments (Section \ref{Models}).

\subsection{Data Privacy Vocabulary (DPV)}
\label{Data Privacy Vocabulary (DPV)}
\begin{figure*}
\centering
\includegraphics[width=14cm]{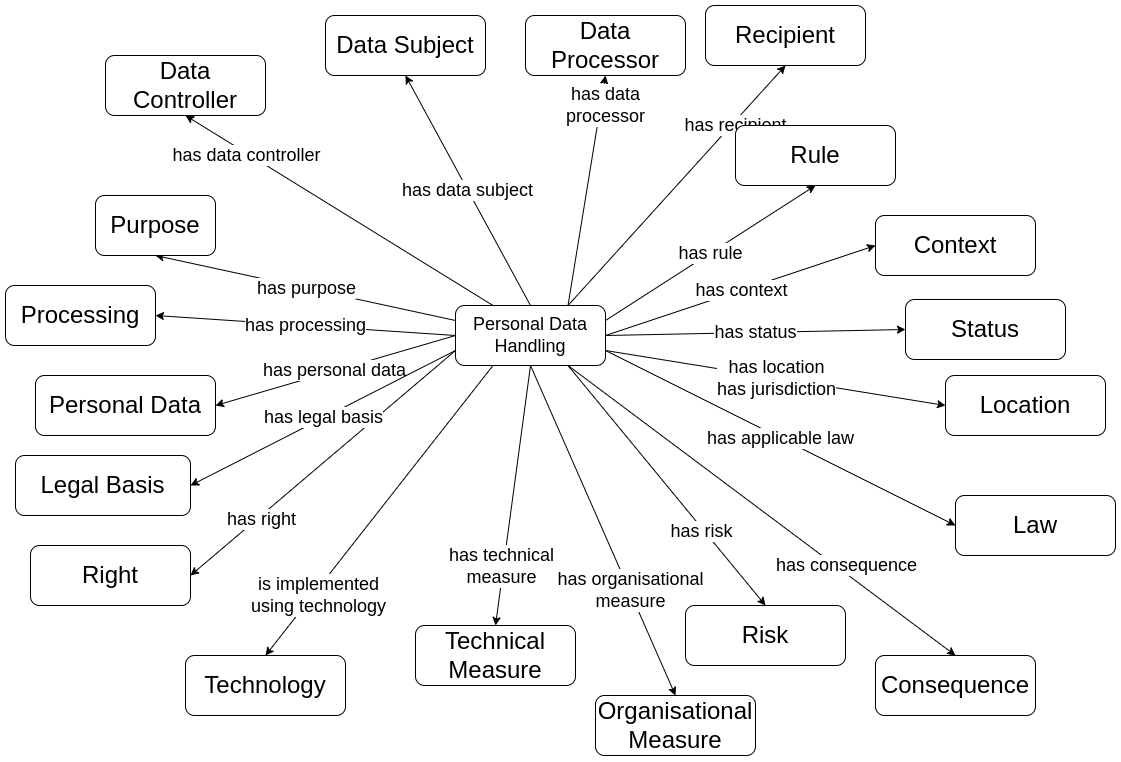}
\caption{Base Vocabulary DPV. The figure can be found in the DPV documentation \cite{DPV-PDOnline}.\\ \href{https://www.w3.org/Consortium/Legal/ipr-notice\#Copyrigh}{Copyright} © 2022 the Contributors to the Data Privacy Vocabulary (DPV) Specification, published by the \href{https://www.w3.org/groups/cg/dpvcg}{Data Privacy Vocabularies and Controls Community Group} under the \href{https://www.w3.org/community/about/agreements/cla/}{W3C Community Contributor License Agreement (CLA).}}. 
\label{fig:Fig.1}
\end{figure*}

As introduced in Section \ref{Related work}, we decided to pay attention to an authoritative resource, the so-called DPV. This resource enables the expression of machine-readable metadata regarding the use and processing of personal data. It provides terms and definitions according to the GDPR and it is divided into classes and properties. The \textit{basic ontology} describes the first-level classes that define a legal policy for the processing of personal data (see Fig.\ref{fig:Fig.1}).

Following the descriptions given in the latest published version of the resource, we are particularly interested in \textit{Personal Data}, for example data directly or indirectly associated with or related to an individual. DPV provides the concept of \textit{Personal Data}, and the relation \textit{has Personal Data} to indicate what categories or instances of personal data are being processed. In particular, \textit{Sensitive Personal Data} is a class for indicating personal data that is considered sensitive in terms of privacy and/or impact, and therefore requires additional considerations and/or protection. The Data Privacy Vocabulary-Personal Data (DPV-PD) extension provides an extended DPV personal data taxonomy, where concepts are structured in a top-down schema based on an opinionated structure contributed by R. Jason Cronk from EnterPrivacy \cite{DPV-PDOnline}.

The DPV-PD presents $206$ Personal Data Categories (PDCs), according to its most recent release (December 05, 2022). Each category is described by a definition and additional information, such as an IRI (Internationalized Resource Identifier), a source and its hierarchical relations.

However, not all categories of the DPV can be explored in the same way. A detailed analysis of the resources led us to identify a narrower set of PDCs to be explored through textual analysis. We divided these categories into 5 different types based on their nature and characteristics that can affect their automatic identification. The subdivision is summarized in Table \ref{tab:Table1}.

\textbf{Macro-categories}: taxonomic organization corresponds to the high-level categories
to which all the more specific PDCs belong. Therefore, their identification was implicit in the identification of nested categories. There are six relevant macro-categories:

\begin{enumerate}
     \item \textit{Historical}: information about historical data related to or relevant to history or past events e.g., \textit{Life History}.
     \item \textit{Financial}: information about finance including monetary characteristics and transactions e.g., \textit{Transactional}, \textit{Ownership}, \textit{Financial Account}.
     \item \textit{Tracking}: information used to track an individual or group e.g. location or email e.g., \textit{Location}, \textit{Device Based}, \textit{Contact}.
     \item \textit{Social}: information about social aspects such as family, public life, or professional networks e.g., \textit{Family}, \textit{Friends}, \textit{Public Life}.
    \item \textit{External} (visible to others): information about external characteristics that can be observed e.g., \textit{Behavioral}, \textit{Physical Trait}, \textit{Physical Characteristic}.
    \item \textit{Internal} (within the person): information about internal characteristics that cannot be seen or observed e.g., \textit{Preference} and \textit{Knowledge Beliefs}.
  \end{enumerate}
\textit{Special Category Personal Data}, cited as a subtype of \textit{Sensitive Personal Data}, is added. The macro-category is based on GDPR Article 9, even if it considers all Sensitive Special Categories whose use is prohibited or regulated with an additional legal basis for justification. Some PDCs include \textit{Health}, \textit{Mental Health}, \textit{Disability}.

Recently, \textit{Household} and \textit{Profile} have also been identified as macro-categories but do not present nested categories.

\textbf{Categories identifiable through textual analysis}: these are categories that can
be frequently expressed through text and whose expressions can be syntactically complex. They are not
alphanumeric sequences or codes that are easily identifiable through regular expressions, but
can be expressed in natural language depending strongly on the combination
of words. Take, for example, the \textit{Age} category,
whose definition is: `\textit{Information about an individual’s age}’. Information about an
individual’s age can be expressed in \textit{n} different ways, such as: `\textit{I’m 17 years old}’ or
`\textit{I was born in 2005}’ or `\textit{In 2010 I was only 5 five years old}’; textual elements are
crucial for its identification. We first investigated these categories.

\textbf{Broad-boundary categories}: these categories can be defined as characterized
by (i) a high degree of vagueness, (ii) a high degree of extension and applicability,
and (iii) whose sensitivity classification is characterized by a high degree of ambiguity. For example, \textit{Intention} which is a sub-category
of \textit{:Preference:Internal} and refers to information about an individual’s
intentions. These categories, owing to their conceptual complexity, have not
been treated as priorities. However, reflections on the future developments of the
work are reserved for them.

\textbf{Uniquely identifiable categories}: these are categories easily identifiable through regular expressions
and fixed sequences, e.g., \textit{Credit Card Number}, \textit{Tax Code}. This type of category (PII) has been heavily extensively in the literature. Tool markets offered by large companies, e.g., Microsoft \cite{Balzer2021} can already be found. It seemed appropriate to focus our analysis on the most
challenging and least explored categories, which could at the same time give us the
possibility of analyzing more complex and context-aware identification techniques.

\textbf{Categories identifiable mainly through non-textual elements}: these depend
completely or largely on non-textual elements and it is therefore difficult, if not impossible, to identify them in this sense. An example may concern the \textit{Fingerprint}
category: `\textit{Information related to an individual’s fingerprint used for biometric purposes}’.

\begin{table*}[t]
\label{tab:Table1}
\centering
\begin{tabular}{p{1cm}p{5cm}p{7.8cm}}
\hline
\multicolumn{1}{c}{{\bf N.}} & \multicolumn{1}{c}{{\bf Type}} & \multicolumn{1}{c}{{\bf Examples}} \\ 
\hline
    6 & Macro-categories & \textsc{Financial, External, Internal, Historical, Social, Tracking} \\
    \hline
    90 & Identifiable through textual analysis & \textsc{Age, Favorite, Health} \\
    \hline
    26 & Broad boundaries categories identifiable through textual analysis & \textsc{Attitude, Interest, Intention}\\
    \hline
    30 & Uniquely identifiable & \textsc{BankAccount, BloodType, CreditCard} \\
    \hline
    54 & Identifiable mainly through non-textual elements & \textsc{Biometric, CallLog, Fingerpint} \\
    \hline
\end{tabular}
\caption{\label{tab:cite-commands} Analysis of the 206 PDCs of the DPV
  }
\end{table*}

Considering the categories identifiable through textual elements, most of the PDCs belong to \textit{Special Data}, \textit{Social} and \textit{External}
macro-categories. In general, the ontological structure arrives at four levels of hierarchy. Some categories, in the analysis and consequent construction of the corpus, were
merged by similarity, e.g., \textit{Physical Characteristic} and \textit{Physical Trait},
or because they are not strictly necessary specifications of a more generic category, e.g.,
\textit{Family} and \textit{Family Structure}.
A list of identified PDCs labels is provided in Table \ref{appendixA}.

\subsection{\spedac: a Sensitive Data Corpus}
\label{Dataset: a sensitive data corpus}
Given the lack of publicly available datasets for sensitive data identification, the first aim of our study is to fill this gap and develop a labeled resource for this task.

Personal data in informal online conversations are the domain context of interest. The TenTen corpus family is a large resource, composed of texts collected from the World Wide Web \cite{Jakubicek2013}. TenTen corpora are available in more than 40 languages. The most recent version of the English TenTen corpus (enTenTen2020) consists of 36 billion words. The texts were downloaded from 2019 to 2021. The sample texts were manually checked and content with poor quality text and the spam was removed. These come from different web domains (the UK
domain .uk, Australian domain .au, Canadian domain .ca, US domain .us, New Zealand
domain .nz, and the EU domain .eu) and different textual genres (news, discussion, blog,
legal) and topics (reference, society, arts, technology, business, sports, science, health,
home, recreation, games); 6.8\% of the corpus comes from English Wikipedia pages. .

For our experiments, we created two different corpora, that were manually labeled by the authors. Both corpora present a sentence-level annotation using INCEpTION as an annotation platform \cite{Klie2018} and  the WebAnno TSV v3.3 annotation format. The datasets are available on a GitHub repository and are shared subject to a declaration of the purposes of use (see Section \ref{Ethical Disclosure}): \url{https://github.com/Gaia-G/SPeDaC-corpora}. 

\paragraph{\textbf{\spedac 1.}} Identification and discrimination of sensitive sentences from non-sensitive sentences. The dataset counts 10,675 sentences and has two target labels:
\begin{enumerate}
    \item \textbf{0 (NON-SENSITIVE)} to indicate sentences without sensitive content.
    \item \textbf{1 (SENSITIVE)} to indicate sentences with sensitive content.
\end{enumerate}

For each fine-grained category (see Table \ref{appendixA}), we have collected sensitive and non-sensitive examples in a balanced manner i.e., considering approximately the same number of examples for each of the two classes. Non-sensitive examples correspond to sentences that contain the same linguistic patterns found in sensitive sentences but in a context that does not confer sensitivity. We can distinguish between two types of linguistic constraints chosen as selection criteria: (i) general constraints and (ii) specific constraints for every PDC. General constraints take into account the importance of the relationship between a PDC and the subject to which it refers. We assume that the identifiable subject (e.g., account, the device used) often corresponds to the person who writes (`I’). The specific linguistic constraints concern multi-word expressions that could better represent every PDC, e.g., the construction `\textit{[have] ... years old}' which recurs to represent the \textsc{AGE} category. As shown in the examples in Table \ref{tab:Table2}, specific constraints are present in sensitive and non-sensitive sentences, whereas the cited general constraint which refers to a first-person subject characterizes only sensitive sentences. Thus, adversarial sentences can better represent ambiguous cases, in which PDCs are present in a non-sensitive context. The adversarial constraints representing the aforementioned cases can consist of citation expressions, e.g., `\textit{[he] [say]}’, `\textit{[article][say]}’, `\textit{[he][state]}’ etc., or expressions concerning the dimension of unreality or supposition, e.g., verbs as `\textit{suppose}', `\textit{imagine}', `\textit{guess}', `\textit{hope}', or adverbs as `\textit{maybe}', or related to the joke e.g., `\textit{just kidding}', `\textit{I [be] joking}'.

\begin{table*}
  \label{tab:Table2}
\centering
\begin{tabular}{p{7.8cm}p{5cm}}
\multicolumn{1}{c}{{\bf Sentence}} & \multicolumn{1}{c}{{\bf Label}} \\ 
    \hline
    hey! I'm 33 years old now. & \textsc{[Age]} \\
    \hline
    The lacquer painting has a history of 80 years old & {[Non-sensitive]} \\
    \hline
   I've suffered depression and other mental probs since my teens & \textsc{[Mental Health]} \\
    \hline
   Mental illness can also be an invisible disability & \textsc{[Non-sensitive]}\\
    \hline
\end{tabular}
\caption{\label{tab:cite-commands}Examples from \spedac 1}
\end{table*}

\paragraph{\textbf{\spedac 2.}}Identification of the PDC macro-categories within sensitive sentences. The 5,133 sentences in the dataset represent the fine-grained PDCs considered in a balanced manner i.e., approximately the same number of examples for each fine-grained category has been taken into account. Specifically, the aim was to collect 100 sentences from each fine-grained PDC.  For some PDCs the retrieval of 100 sensitive sentences was difficult and they are therefore less represented in the corpus, e.g., \textit{Criminal}, \textit{Criminal Conviction}, \textit{Criminal Charge}, \textit{Disciplinary Action}, \textit{Income Bracket}, \textit{Privacy Preference}, \textit{Professional Evaluation}, \textit{Professional Interview}, \textit{Salary}, \textit{Skin Tone}.

For every sentence its macro-category has been retrieved, presenting totally 5 different labels, which are the following: 

\small
\begin{enumerate}
    \item Special Category Data
    \item Financial and Tracking
    \item Social
    \item Internal
    \item External
\end{enumerate} 

\normalsize
The category \textsc{Historical} has been excluded because of its inconsistency (it is a superclass only of \textit{Life History} PDC, which is a broad-boundary category).

The percentage of representation of the macro categories in the corpus, which depends on the number of specific categories included, is presented in Table \ref{appendixA}.

\paragraph{\textbf{Inter-annotator agreement}} To measure the goodness of our annotations, we asked a group of linguists to annotate a sample from each corpus. The basis given to them for annotation was the taxonomy of DPV-PD.
\begin{enumerate}
    \item \spedac 1: we asked 4 annotators to binary classify 100 sentences as sensitive or non-sensitive. Giving the taxonomy as a reference, it was specified not to mark as sensitive only the sentences containing PII but to follow a more extensive definition of personal information that takes into consideration all the PDCs listed in the provided taxonomy;
    \item \spedac 2: we asked 3 annotators to classify 150 sentences over the 5 macro-categories of the PDCs. In addition to the taxonomy, a detailed definition of the 5 macro-categories was provided, with examples of PDCs included in each group; 
    \item \spedac 3: because the specific PDCs are numerous, we have limited the task to the validation of our first labeling on 50 sentences. We received contribution from 4 annotators. They were asked to compare the specific PDCs with which they found the sentences labeled with the definition given in the DPV-PD.
\end{enumerate}

Sentences were randomly selected, balancing the number of different labels on \spedac 1 and \spedac 2.

We measured the score agreement by aggregating the original annotation with the others using the Krippendorff’s alpha ($\alpha$)  coefficient \cite{Hayes2007}. Krippendorff's $\alpha$ expresses the score in terms of disagreement and is recommended if there are 3 or more annotators, attenuating the statistical effects of samples low-size datasets and ignores missing data that may be present in collaborative work. Values range from 0 to 1, where 0 indicates perfect disagreement and 1 indicates perfect agreement.  ($\alpha$) $\geq$ .800  is usually considered a high agreement, while an acceptable agreement is considered in Krippendorff \cite{Krippendorff2004} as .667 $\geq$ ($\alpha$) $\geq$ .800, even if the various proposals of the scholars highlight the arbitrary character of the reference thresholds \cite{Gagliardi2018}.

The Krippendorff's ($\alpha$) is 0.73 for \spedac 1, 0.82 for \spedac 2, and 0.87 for \spedac 3 respectively.
The score obtained by comparing the gold annotation with the annotation or validation of each annotator was also measured. The results are summarized in table \ref{tab:Tabled1}. It might be unusual to see a higher percentage of agreement in the second task, where there are more labels. In \spedac 1 the sentences that reported a high rate of disagreement are mostly (i) ambiguous sentences, in which potentially sensitive personal data is expressed, as well as the
relationship with a subject, but appear within a non-sensitive context (e.g., a fictitious example to explain a concept); (ii) sentences in which potentially sensitive personal data
appears but the subject is not uniquely identifiable (often an unspecified group of people); (iii) sensitive sentences presenting specific personal data not identified by annotators,
e.g., \textit{House Owned}. On the other hand, despite obtaining an ‘almost perfect’ agreement score, \spedac 2 and \spedac 3 can sometimes present sentences that are potentially multi-labeled.

\paragraph{Dataset split} Each dataset was randomly divided into three parts for the experimental process: 70\% training set, 10\% development set, and 20\% test set (see Table \ref{tab:Table3}).

\begin{table}[t]
\label{tab:Table3} 
\begin{center}
\begin{tabular}{|l|l|l|l|}
\hline 
& \bf \spedac 1 & \bf \spedac 2 & \bf \spedac 3 \\  \hline

    \bf Training set & 7611 & 3695 & 3893 \\
    \bf Validation set & 846 & 411 & 556 \\
    \bf Test set & 2218 & 1027 & 1112 \\
\hline
\hline
\end{tabular}
\end{center}
\caption{Size of Dataset used for experiment}
\end{table}

\begin{table}[t]
\label{tab:Tabled1} 
\begin{center}
\begin{tabular}{|l|l|l|l|l|}
\hline  & \bf Ann. 1 & \bf Ann. 2 & \bf Ann. 3 & \bf Ann. 4 \\ \hline

    \bf \spedac 1 & 0.84 & 0.82 & 0.68 & 0.68 \\
    \bf \spedac 2 & 0.82 & 0.84 & 0.92 & / \\
    \bf \spedac 3 & 0.94 & 0.90 & 0.84 & 0.88 \\
\hline
\hline
\end{tabular}
\end{center}
\caption{Krippendorff's (α) between gold and single annotation}
\end{table}

The distribution of labels in the training, validation and test sets of \spedac 1 and \spedac 2 can be observed in Table \ref{tab:Table3.1} and Table \ref{tab:Table3.2}.

\begin{table*} [t]
\label{tab:Table3.1}
\centering
    \begin{tabular}{@{} cccc||cc||cc}
    \cline{2-8}
        & & Train  & \% Train &  Val   &  \% Val  & Test  & \% Test     \\ [0.2ex] % inserts table 

        \cline{2-8}
        & \textbf{Non-sens} & 3790 & 49.80\% & 405 & 47.87\% & 1086 & 48.96\% \\ 
        & \textbf{Sens} & 3821 & 50.20\% & 441 & 52.13\% & 1132 & 51.04\% \\ 
        %\cmidrule[1pt]{2-12}
           \cline{2-8}
    \end{tabular}\\
    \caption{Label distribution in \spedac 1}
    \end{table*}

\begin{table*} [t]
\label{tab:Table3.2}
\centering
    \begin{tabular}{@{} cccc||cc||cc}
    \cline{2-8}
        & & Train  & \% Train &  Val   &  \% Val  & Test  & \% Test     \\ [0.2ex] % inserts table 

        \cline{2-8}
        & \textbf{Special Data} & 979 & 26.49\% & 103 & 25.06\% & 274 & 26.68\% \\ 
        & \textbf{Financial and Tracking} & 468 & 12.67\% & 59 & 14.36\% & 122 & 11.88\% \\ 
        & \textbf{Social} & 1100 & 29.77\% & 137 & 33.33\% & 328 & 31.94\% \\ 
        & \textbf{Internal} & 321 & 8.69\% & 30 & 7.30\% & 83 & 8.08\% \\ 
        & \textbf{External} & 827 & 22.38\% & 82 & 19.95\% & 220 & 21.42\% \\ 
        %\cmidrule[1pt]{2-12}
           \cline{2-8}
    \end{tabular}\\
        \caption{Label distribution in \spedac 2}
    \end{table*}

\subsection{Models}
\label{Models}
We dedicate this paragraph to the description of the computational models used to conduct classification experiments on the different tasks offered by \spedac.

\textbf{Baseline.}
The baseline was calculated using the Zero Rate (ZeroR) classifier.
This method draws the most-frequent baseline by roughly classifying all instances as corresponding to the most frequent class.

\textbf{k-Nearest Neighbors ($k$-NN).}
$k$-NN is an algorithm used both for classification and for
regression, which is based on the similar characteristics of neighboring features \cite{Wang2012}. The $k$-NN classifier uses instance-based learning, it does not build a general internal model, but stores instances of the training data. An instance is classified based on the plurality vote of its closest neighbors. The data class that has the greatest number of representatives within the closest neighbors to the instance is the predicted class. The number of neighbors to be considered is a parameter of the model to be established (k). In particular, for binary and multiclass classification, the number of neighbors should be odd. We trained the model implemented in sklearn \cite{Pedregosa2011}: KNeighborsClassifier\footnote{The model implemented can be found here: \url{https://scikit-learn.org/stable/modules/generated/sklearn.neighbors.KNeighborsClassifier.html} (last access December 05, 2022)}, where the optimal choice of the value k is highly data-dependent (generally, a larger k can reduce the noise, but makes the classification boundaries less distinct). Time complexity of the model is defined - following the Big O notation \cite{Kearns1990} - by the product of \textit{k}=number of neighbors;\textit{d}=number of data points; and \textit{n}=number of neurons/data dimensionality. The time complexity of the models used are summarized in Table \ref{tab:Table3C}.

\noindent\textbf{Support Vector Machines (SVMs).}
Support Vector Machines (SVMs) are another classic algorithm \cite{Cortes1995} capable of building both binary and multiclass classifiers. SVMs use tagged data to define a hyperplane in which they map training examples, in an attempt to maximize the gap between categories. New examples are classified based on where they are mapped. For multiclass classification, the same principle is used after breaking down the classification problem into smaller subproblems, all of which are binary classification problems. A LIBSVM linear model was used from sklearn \cite{Pedregosa2011} for our experiments\footnote{The model implemented can be found here: \url{https://scikit-learn.org/stable/modules/generated/sklearn.svm.SVC.html} (last access December 05, 2022)}. The model is recommended for sets that are not too wide: the fit time scales at least quadratically with the number of samples, and therefore excludes its use on large datasets. The time complexity of SVMs is calculated with an exponent of 3 \cite{Abdiansah2015}.

\noindent\textbf{Logistic Regression.}
Logistic regression (LR) is a regression model implemented for binary and multiclass classification problems \cite{Bisong2019}.
The model establishes the probability of identifying the value of the dependent variable by analyzing the attributes of the input and processing a weight distribution. The probability of belonging to a sample was calculated for each class. We used the LR model implemented in sklearn \cite{Pedregosa2011} for the experiments\footnote{The model implemented can be found here: \url{https://scikit-learn.org/stable/modules/generated/sklearn.linear_model.LogisticRegression.html} (last access December 05, 2022)}. This implementation can fit binary, One-vs-Rest, or multinomial logistic regression with optional penalty terms (\textit{l1,l2}),  or Elastic-Net regularization (by default). Time complexity is a product of data dimensionality and number of data inputs.

\noindent\textbf{Transformer-based Language Models: RoBERTa and DeBERTa.}
We adopted the RoBERTa (Robustly Optimized BERT Pre-training Approach) and DeBERTa (Decoding-enhanced BERT with disentangled attention) transformer architecture \cite{Liu2019,He2021}. 
RoBERTa has been proven to perform well in different NLP tasks, including classification \cite{Briskilal2022,Qiu2020,Bilal2022}. DeBERTa models seem to perform consistently better on a wide range of NLP tasks even if trained on half of the training data \cite{He2021}. RoBERTa and DeBERTa are both an extension of the Bidirectional Encoder Representations from Transformers (BERT) \cite{Devlin2019}. BERT uses two bidirectional training strategies: the Masked Language Model (MLM), which deals with the relationship between words and the Next Predictive Sentence (NPS) to predict the relationship between sentences. BERT's architecture is composed of a tokenizer (WordPiece) and a large stack of transformers, which is provided with the input for training. The BERT-Base model consists of a 12-layer transformer, whereas the BERT-Large of 24-layer. RoBERTa has almost the same architecture as BERT model, but uses a byte version of Byte-Pair Encoding (BPE) as a tokenizer and is pretrained with the MLM task (without the NPS task). It optimizes some hyperparameters for BERT, e.g., longer training time, larger training data, larger batch size, larger vocabulary size, and dynamic masking.
DeBERTa improves the BERT and RoBERTa models adding two novel techniques. First, a disentangled attention mechanism uses two vectors to encode and separate the content and position of a word. Second, an enhanced mask decoder is able to predict both relative and absolute position of words, while the previous models took into account only one of them.

We used the RoBERTa-base and DeBERTa-base models with pre-trained weights \cite{RoBERTaHuggingface,DeBERTaHuggingface} and 768 hidden dimensions. Time complexity is a product between \textit{n} with an exponent of 2 and \textit{d} considered per layer \cite{Vaswani2017}. The additional computational complexity of DeBERTa is $O (k * n * d)$ due to the calculation of the additional position-to-content and content-to-position attention scores. This increases
the computational cost of RoBERTa by 30\% \cite{He2021}.

\noindent\textbf{Sentence-transformers Language Models: LaBSE.}
The architecture of a sentence-transformer model is made of two main layers. The first layer is a transformer model with a  fixed lenght of 768 dimensions that outputs contextualized word embeddings for all input tokens. The second layer is a pooling layer that average the embeddings generated by the model giving a fixed lenght vector \cite{Reimers2019}.
LaBSE (Language-agnostic BERT sentence embedding) \cite{Feng2020} is a multilingual sentence embedding model for more than 109 languages, originally trained and optimized in order to produce similar representations exclusively for bilingual sentence pairs that are translations of each other. Thanks to its specific training, the model achieves state-of-the-art performance on bilingual retrieval/mining tasks. Multilingual sentence embedding models produce representations that are suitable to be compared with simple cosine similarity also on the same language \cite{Feng2020,tripodi-etal-2022-evaluating}. The study aims to investigate the use of a LaBSE model on classification task.

The used encoder architecture follows the BERT-Base
model, with 12 hidden-layers and 768 per-position hidden units. Sentence embeddings are extracted from the last transformer block \cite{LaBmodel}.

\begin{table}[t]
\label{tab:Table3C} 
\begin{center}
\begin{tabular}{|l|l|l|}
\hline  \bf Model & \bf Time complexity & \bf Layers and Parameters \\  \hline
\bf k-NN & $O (k*n*d)$ & /  \\
\bf SVMs & $O (n^3)$ & / \\
\bf LR & $O (n * d)$ & / \\
\bf LaBSE & $O (n^2*d)$
                per layer & 12-layers, 471M\\
\bf RoBERTa & $O (n^2*d)$
                per layer & 12-layers, 125M\\
\bf DeBERTa & $O (n^2*d)$
                per layer & 12-layers, 98M\\
                
    \hline
\hline
\end{tabular}
\end{center}
\caption{Time complexity of the models following Big O notation \cite{Kearns1990}}
\end{table}

\section{Experimental Setup}
\label{Experimental process}
The datasets created and described in Section \ref{Dataset: a sensitive data corpus} were used first for the experiments conducted with the transformer models and then, for comparative purposes, with the other models.

\subsection{Experiment 1.} \textit{Dataset.} Identification of sensitive sentences and exclusion of non-sensitive sentences. In particular, as described above, we built an adversarial dataset of sentences with non-sensitive content that is particularly competitive with the sensitive content dataset. The sentences in both datasets contain the same linguistic patterns; what differentiates a sensitive sentence from a non-sensitive one is the context in which it occurs. The same datasets were used to perform all the experiments. The subdivision, as described in Tables \ref{tab:Table3}, \ref{tab:Table3.1} occurred randomly only once and the derived datasets were used to train and test all models.

\noindent\textit{Preprocessing, features and parameters.} First, the data were preprocessed and cleaned. The preprocessing process includes tokenization of sentences, lemmatization, conversion of each token into a lower case, removal of spaces, stop words and punctuation.
Feature extraction on the training set was performed using the scikit learn feature extraction from the text modules. The features are not domain dependent but English language dependent, and are the following:
\begin{itemize}
    \item whether a token starts and ends a sentence;
    \item the length of the sentences in tokens;
    \item Bag-Of-Words (BOW) vectors (ngram range=1,1) using the SPaCy CounVectorizer function \cite{Pedregosa2011}. 
\end{itemize}

Preprocessing and feature extraction using SpaCY are common for all three classic machine learning models implemented (kNN, SVM, and LR).

The model parameters were set up and tuned on the \spedac validation set as follows:

\begin{itemize}
    \item for the $k$-NN model, we considered the 3 closest neighbors (k=3);
    \item for the SVM model, we used default parameters to set up a linear kernel;
    \item for the LR model, default parameters have been used;
    \item for the transformer models, we set a stack with dropout level of 0.3, and a randomly initialized linear transformation level above the model. The maximum sequence length was set to 256, and the training lot size was set to 8. For the model optimization, we used the AdamW optimizer \cite{Loshchilov2018} with a learning rate of 1\textit{e}-5. The performance was evaluated based on the loss of the binary cross-entropy. After 3 epochs, the model reports a training accuracy epoch beyond 0.90 on the validation set.
\end{itemize}

\subsection{Experiment 2.} 
\begin{figure*}[t]
\centering
\includegraphics[width=12cm]{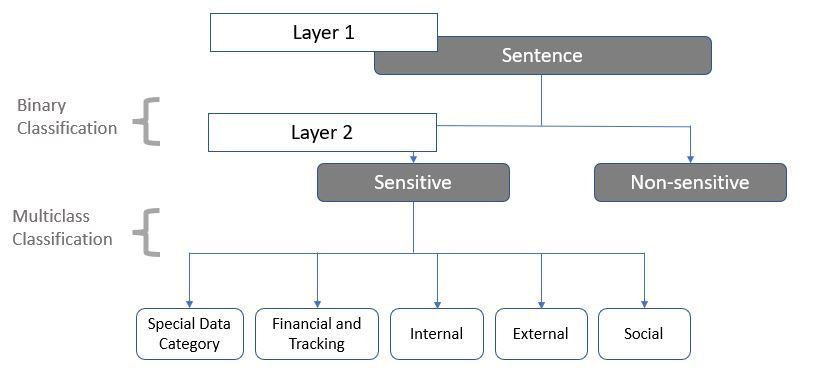}
\caption{Flow of Sensitive Detection model} 
\label{fig:Fig.4}
\end{figure*}

\textit{Dataset.} Identification of which type of sensitive data the sentence presents, related to its own macro-category. Once the sensitive sentences have been identified (layer 1, Fig.\ref{fig:Fig.4}), they are analyzed by the multiclass model, which labels them according to the 5 macro-categories on which they are trained 
(layer 2, Fig.\ref{fig:Fig.4}).

Even in this case, the same datasets were used to run all experiments and their subdivisions are described in Tables \ref{tab:Table3}, \ref{tab:Table3.2}.

\noindent\textit{Preprocessing, features and parameters.} 
The preprocessing process and feature extraction is the same of the first experiment.

The parameters of the models, set up and tuned on the \spedac validation set, are as follows:

\begin{itemize}
    \item for the $k$-NN model, we considered the 3 closest neighbors (k=3);
    \item for the SVM model, the multiclass classification strategy used follows the One-vs-One (OvO) scheme, which involves breaking down the multiclass classification into a binary classification problem for each pair of classes;
    \item for the LR model, this case, for the multiclass classification we used the One-vs-Rest (OvR) scheme, which divides multiclass classification into a binary classification problem by class;
    \item for the transformer models, the setting is the same as for \spedac 1, and likewise reports a training accuracy epoch beyond 0.90 on the validation set.
\end{itemize}

\subsection{Experiment 3.}

\textit{Dataset.} Identification of the type of fine-grained PDC in a sentence. This involves a multiclass classification task with 61 labels and a small amount of training data for each PDCs.

\noindent\textit{Preprocessing, features and parameters.}
The models used were the same as those in the second experiment with the following differences:
\begin{itemize}
    \item for the baseline of \spedac 3, the 61 labels were traced to the macro-category and the most-frequent baseline was calculated by tracing all the test sentences to the most frequent macro-category;
    \item for the $k$-NN model, 5 closest neighbors has been used (k=5);
    \item to improve the LR results, a liblinear solver with penalty \textit{l1} was applied;
    \item to improve the transformer models results, a category regularization with a label smoothing technique was introduced \cite{Muller2019} and the number of epochs in training was increased to 15.
\end{itemize}

\section{Results}
\label{Results}

\begin{table}[t]
\label{tab:Table4} 
\begin{center}
\begin{tabular}{|l|l|l|}
\hline  & \bf \spedac 1 & \bf \spedac 2 \\  \hline
\bf Baseline & 51.04\% & 31.93\% \\
\bf k-NN & 68.62\% & 63.78\% \\
\bf SVM & 93.15\% & 92.30\%  \\
\bf LR & 92.60\% & 92.50\% \\
\bf LABSE & 98.15\% & 94.84\% \\
\bf RoBERTa & \textbf{98.20\%} & 94.94\% \\
\bf DeBERTa & 98.11\% & \textbf{95.81\%} \\
    \hline
\hline
\end{tabular}
\end{center}
\caption{Accuracy results on \spedac 1 and \spedac 2}
\end{table}

The model predictions were evaluated in terms of accuracy.

\textit{Experiment 1.}
The results of the first and second experiments on \spedac 1 are listed in Table \ref{tab:Table4}.
As can be seen, RoBERTa reports the best results compared with the other models for the binary classification task for sensitive and non-sensitive sentence identification even if DeBERTa and LaBSE both report very high results as well. \spedac 1, as described in Section \ref{Dataset: a sensitive data corpus}, is composed of sensitive and non-sensitive sentences that have the same linguistic patterns, which acquire sensitivity or not depending on the context. If the discriminant of sensitive and non-sensitive sentences in the dataset often consists of contextual elements, given the occurrence of the same linguistic patterns, the transformer  context-aware models turns out to be the most suitable for the task.

\textit{Experiment 2.} The results of the first and second experiments on \spedac 2 are listed in Table \ref{tab:Table4}. In the multi-class classification of \spedac 2, where the problem of ambiguity is less evident, the results obtained with the other models are more promising. The DeBERTa model outperforms the other models in all cases, and the RoBERTa model surpasses the LR performance by 2.44\%.

It is interesting to observe how LaBSE, a very promising model for multilingual sentence similarity, does not achieve the best results for the classification task when compared to the other transformer models. Probably because it was trained to detect similar sentences in different languages.

\textit{Experiment 3.} The models performances of the third experiment on \spedac 3 are presented in Table \ref{tab:Table4.1}. 
 The results, which differ significantly between the models in terms of percentage accuracy offer valid results for a benchmark on the \spedac 3.

\begin{table}[t]
\label{tab:Table4.1} 
\begin{center}
\begin{tabular}{|l|l|}
\hline  & \bf \spedac 3 \\  \hline
\bf Baseline & 32.25\% \\
\bf k-NN & 35.30\% \\
\bf SVM & 57.59\% \\
\bf LR & 75.74\% \\
\bf LABSE & 77.09\% \\
\bf RoBERTa & 77.18\% \\
\bf DeBERTa & \textbf{77.63\%} \\
    \hline
\hline
\end{tabular}
\end{center}
\caption{Accuracy results on \spedac 3: a new benchmark}
\end{table}

\subsection{Results Analysis}
\label{Results analysis}

\begin{figure*} [t]
\centering
\includegraphics[width=14cm]{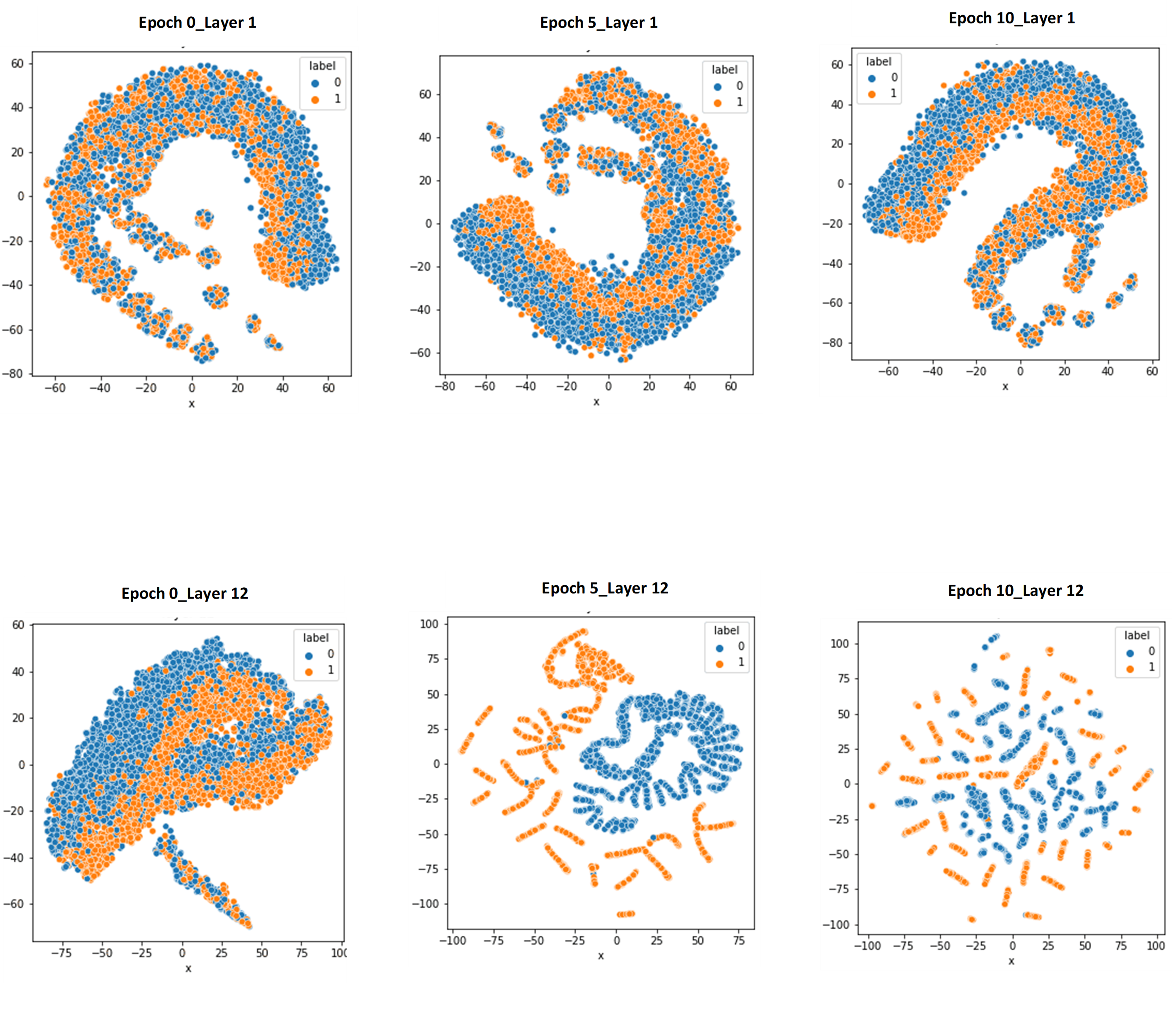}
\caption{RoBERTa embeddings t-SNE visualization during 2-class fine-tuning (perplexity=30)}. 
\label{fig:Fig.4.1}
\end{figure*}

\begin{figure*} [t]
\centering
\includegraphics[width=14cm]{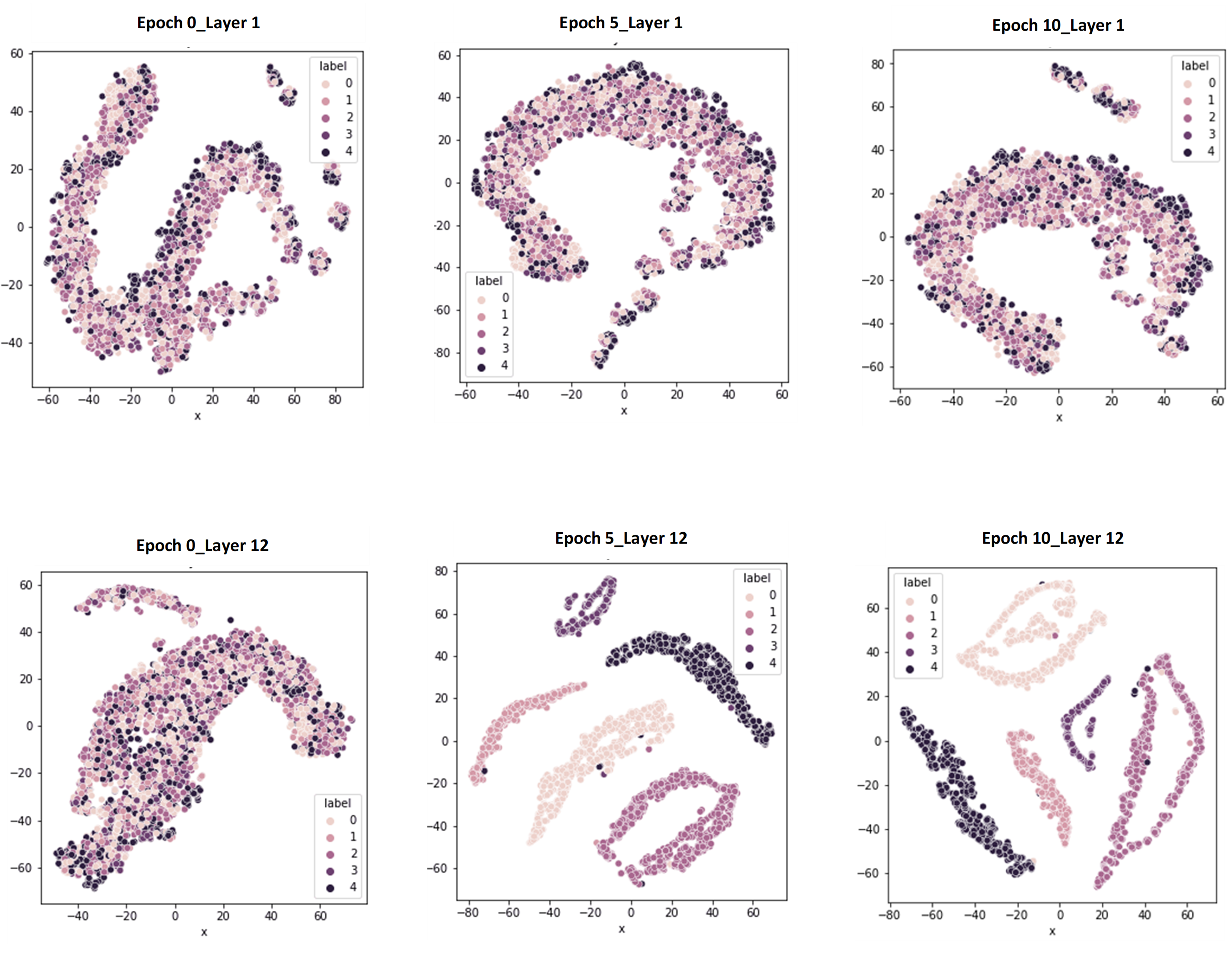}
\caption{RoBERTa embeddings t-SNE visualization during 5-class fine-tuning (perplexity=30)}. 
\label{fig:Fig.4.2}
\end{figure*}

Fig. \ref{fig:Fig.4.1} and Fig. \ref{fig:Fig.4.2} show a t-SNE visualization\cite{Maaten2008} of the RoBERTa embeddings during the fine-tuning on training data. The first and last hidden layers of the transformer network are reported. During the validation stage the weights of the model are not updated. From visualizations, it can be seen that for both tasks, already after epoch 5, the embeddings are distinctly clustered.

To better understand the behavior of the model on \spedac 1 and \spedac 2, we report the results in terms of the accuracy for each classification category taking RoBERTa and DeBERTa as the transformer models that obtain the best results and SVMs of the ML comparison methods (see Table \ref{tab:Table5},\ref{tab:Table6}). By analyzing the errors through confusion matrices, we see how the RoBERTa and DeBERTa models obtain the best performance for each category without significant differences; there are no particularly critical categories to classify.

\begin{table*}
\centering
\label{tab:Table5}
\renewcommand\arraystretch{1.6}
\settowidth\rotheadsize{\theadfont Actual Class}
    \begin{tabular}{@{} cccc||cc||cc}
            & & \multicolumn{4}{c}{Predicted Class} \\[2ex]
        \cline{2-8}
        & &  RoBERTa  &    & DeBERTA & & SVM   &      \\ [0.2ex] % inserts table 
        \cline{2-8}
        &              & \textbf{Non-sens} & \textbf{Sens} & \textbf{Non-sens} & \textbf{Sens} & \textbf{Non-sens} & \textbf{Sens}\\ [0.5ex] % inserts table 
        \cline{2-8}
    \multirow{2}{*}[9ex]{\rothead{Actual Class}}
        & \textbf{Non-sens} & \textbf{1102} & 30 & \textbf{1104} & 28 & \textbf{1036} & 96 \\ 
        & \textbf{Sens} & 10 & \textbf{1076} & 14 & \bf{1072} & 56 & \textbf{1030} \\ 
        %\cmidrule[1pt]{2-12}
           \cline{2-8}
    \end{tabular}\\
\caption{Confusion Matrix sensitiveness classification on \spedac 1}
\end{table*}

\begin{table}\small
\label{tab:Table6}
\renewcommand\arraystretch{1.2}
\settowidth\rotheadsize{\theadfont Actual Class}
    \begin{tabular}{@{} ccccccc||ccccc||ccccc}
            & & \multicolumn{15}{c}{Predicted Class} \\[2ex]
        \cline{2-17}
        &  &  & RoBERTa  &  & &  & & & DeBERTa &  & &  & & & SVM   &      \\ [0.5ex] % inserts table 
        \cline{2-17}
        &              & \textbf{Spec} & \textbf{Fin} & \textbf{Soc} & \textbf{Int} & \textbf{Ext}
         & \textbf{Spec} & \textbf{Fin} & \textbf{Soc} & \textbf{Int} & \textbf{Ext}
         & \textbf{Spec} & \textbf{Fin} & \textbf{Soc} & \textbf{Int} & \textbf{Ext} \\
        \cline{2-17}
    \multirow{5}{*}[1ex]{\rothead {Actual Class}}
        & \textbf{Spec} & \textbf{258} & 0 & 10 & 0 & 6 & \textbf{259} & 0 & 4 & 1 & 10 &  \textbf{248} & 1 & 14 & 0 & 11 \\
        & \textbf{Fin} & 1 & \textbf{115} & 6 & 0 & 0  & 0 & \textbf{120} & 1 & 0 & 1 & 1 & \textbf{114} & 6 & 0 & 1 \\
        & \textbf{Soc}  & 4 & 2 & \textbf{319} & 2 & 1 &  8 & 3 & \textbf{312} & 3 & 2 & 7 & 7 & \textbf{305} & 1 & 8 \\
        & \textbf{Int} & 1 & 0 & 1 & \textbf{81} & 0 & 1 & 0 & 0 & \textbf{82} & 0 & 0 & 0 & 0 & \textbf{81} & 2 \\
        & \textbf{Ext} & 7 & 1 & 10 & 0 & \textbf{202} & 6 & 1 & 2 & 0 & \textbf{212} & 12 & 2 & 6 & 0 & \textbf{200} \\
        %\cmidrule[1pt]{2-12}
           \cline{2-17}
    \end{tabular}\\
    \caption{Confusion Matrix macro-categories classification on \spedac 2}
\end{table}
\noindent

However, it should be noted that, in terms of time complexity (Table \ref{tab:Table3C}), the ML models report significantly lower values than the transformer ones.

Concerning the specific experiments we can make the following considerations:

\textit{Experiment 1.} The models mostly failed to identify non-sensitive sentences, although RoBERTa and DeBERTa are considerably more accurate. By analyzing errors, many sentences are misidentified as sensitive presumably because of the high rate of ambiguity they present. Errors are caused by the presence of expressions and keywords related to health or profession and the model in these cases is unable to discriminate assumptions or hopes useful to exclude the sensitivity of the sentence e.g., `\textit{I had great hopes of being an air hostess so that i could travel to so many places than
I heard about a plane crushed and that kind of threw me off the idea}'  However, it is important to note that this is not a systematic error: RoBERTa at the same time classifies as non-sensitive sentences where the profession of the subject is only a guess e.g., `\textit{I 'm supposed to be a movie critic, and yet I keep hearing about these great new movies I've never seen.}'. This leads us to consider that cases of ambiguity can be addressed by adding training sentences to represent them.

\begin{table*} [t]
\label{tab:Table11}
\centering
    \begin{tabular}{|l|l|l|l|l|l}
\hline  \bf Target & \bf Pred & \bf \% Error RoB & \bf \% Error DeB & \bf \% Error LR \\  \hline
 Ethnicity and Ethnic Origin & Skin Color & 21.40\% & 21.40\% & 21.40\% \\
 Family Health History & Drug Test Result & 20\% & 20\% & 20\% \\
 Favorite Food & Favorite & /  & / & 28.50\% \\
 Location & Country &  33.30\% & 20\% & 40\% \\
 Health History & Health &  24\% & / & / \\
 Mental Health & Health & 20\% & 26.70\% & 20\% \\
 Physical Characteristic and Trait & Hair Color & 36.80\% & 26\% & 31.50\% \\
 Professional Evaluation & Reference & 25\% & / & 50\%
 \\
 Reference & Employment History & 20\% & / & / \\
 Reference & Professional Interview & / & 20\% & 20\% \\
 Salary & Parent & 25\% & 25\% & 25\% \\
 Salary & Credit & 25\% & / & / \\
 Salary & Family and Family Structure & / & / & 25\% \\
 School & Professional Certification & 31.20\% & / & / \\
 Sexual & Proclivitie & 31.80\% & 31.80\% & / \\
 Sexual History & Sexual & / & / & 28\% \\
 Work History & Employment History & 47.60\% & 38\% & 61.90\% \\
    \hline
\hline
\end{tabular}
 \caption{\% Errors $\geq$ 20\% in \spedac 3 (RoBERTa, DeBERTa and LR models). When '/' appears, it means a \% of error $<$ 20.}
    \end{table*}

\textit{Experiment 2.} Contrary to what one might assume, the category with fewer training examples (\textit{Internal}) achieved a high accuracy score. Indeed, the macro category has fewer examples, but at the same time has fewer specific categories of personal data that represent it. All the specific PDCs belonging to the macro PDC \textit{Internal} refer to personal preferences (\textit{Preference}, \textit{Favorite Color}, \textit{Favorite Music}) and are therefore well identified by the model. RoBERTa mainly mistakes the macro PDC \textit{External} for the \textit{Social} and the category is generally confused with the \textit{Special Data} category. Furthermore, it can be seen that the models confuse some sentences classified as \textit{Special Data} or \textit{External} with the \textit{Social} category. This can be explained by the fact that some sentences contained more than one PDC. Therefore, they would need double-label and classification.

\textit{Experiment 3.} As for the \spedac 2 experiment, the models that have achieved the highest performance were the RoBERTa and DeBERTa transformer models and the LR-based models.

By conducting an error analysis on the predictions of the models, we identified systematic confusions between targets and predictions, highlighting the errors that exceeded 20\% (see Table \ref{tab:Table11}). The confusing labels often belong to the same macro-category and present similarities in terms of keywords and linguistic patterns.

Another significant problem that emerges from the error analysis concerns sentences that contain more than one sensitive data item which would require multi-category labeling. Example: `\textit{Nancy and I were married in 1977 and we lived for nearly 30 years in the Duveneck school area}' is a sentence that reveals sensitive information that can be traced back to two categories: \textit{Marital Status} and \textit{Location}. In future works, it is expected that this problem will be solved by a span-based labelling of \spedac.

\section{Conclusion and Future Work}
\label{Conclusion and Future Work}
In this study, we investigated the task of automatic sensitive data identification and classification, based on our work on personal data categories, which has not been explored in the literature. To do this, we created labeled datasets. The \spedac corpora were evaluated by comparing of machine learning algorithms, including the transformer models, with which we achieved the best results. An accuracy of over 90\% was achieved in the classification of sensitive and non-sensitive sentences (\spedac 1) and in the discrimination of the 5 macro-categories of personal data. Lower results ($<$ 80\% acc.) are achieved in the 61-class classification of \spedac 3. This dataset can be used as a valid benchmark for future studies.

First, the most important goal achieved in this work concerns the creation of the \spedac labeled datasets for the task of automatic identification of personal data, based on the taxonomy of the DPV. The datasets constitutes an available resource and a benchmark for the task, which is currently not present in the literature. A future work foresees the expansion of the \spedac corpora both quantitatively and multilingual. In particular, we would like to consider the Italian language. Therefore, as anticipated and based on the error analyses conducted in the experiments, it would be very useful to label \spedac at a finer level than sentence-based one, labeling of multiple PDCs on each single sentence. We assume token-level labeling following the BIO encoding format.

Second, to evaluate \spedac, we explored a model based on deep learning for the identification of sentences with sensitive content and the classification of the personal data macro-categories present in them. The hypothesis that pre-trained transformer networks based on multi-head attention modules can perform classification tasks whose labels are highly context-dependent has been confirmed by the results. Indeed, binary and 5-label classification tasks conducted on the BERT extension, DeBERTa, report extremely high accuracy results and appear to be the best especially when compared to different automatic learning models ($k$-NN, SVM, and Logistic Regression).

However, the deep learning approach does not seem to achieve excellent results when there are few training data and many classification labels, as in the case of \spedac 3, although model adaptation techniques (e.g., label smoothing) can improve them.  Combining a logical-symbolic approach that requires little or no training data could be an interesting solution to explore \cite{Gambarelli2022}. 

In any case, the comparison with the state-of-the-art when implementing different identification techniques is always very difficult, because of the lack of shared resources and benchmarks. The \spedac contributes in this sense. The datasets can be shared under an ethical disclosure agreement and used to evaluate other identification and classification models for PDCs.

To conclude, the SID task we have addressed, which - as aforementioned - is a subpart of the DLD,  helps to improve the DLP systems. The resource and the results intercept an industrial interest. Future works could also explore and test the model to search for and identify sensitive information in structured data. Finally \spedac could be extended to identify other sensitive data categories at high risk of DLD e.g., passwords left in scripts and software codes.

\section{Ethical Disclosure}
\label{Ethical Disclosure}
The automatic processing of sensitive data implies a necessary reflection on the ethical aspects and improper uses that derived from this type of research \cite{Suster2017}, \cite{Weidinger2021}. The created dataset presents publicly available texts, labeled by categories of sensitive data but in no way attributable to identifiable subjects. This dataset simulates the contexts of sensitivity, but is not actually sensitive. Nevertheless, the trained model can certainly be used for malicious purposes, in contrast to what we pursue. To avoid this possibility, we have bound the download of \spedac to the prior signing of an agreement by the user that establishes ethical research purposes.
\clearpage

\begin{appendix} 

\centering
\begin{xltabular}{\textwidth}{XXl}

\label{appendixA}
    \textbf{Label} & \textbf{PDCs} & \textbf{\% dataset} \\
    \hline
    [Age] & \texttt{AGE, AGE EXACT, AGE RANGE, BIRTH DATE, BIRTH PLACE} & 1.87\% \\
    \hline
    [Apartment Owned] & \texttt{APARTMENT OWNED} & 1.74\% \\
    \hline
    [CarOwned] & \texttt{CAR OWNED} & 1.51\% \\
    \hline
    [Country] & \texttt{COUNTRY} & 1.71\%  \\
    \hline
    [Credit] & \texttt{CREDIT} & 1.73\% \\
    \hline
    [Criminal] & \texttt{CRIMINAL,CRIMINAL CHARGE, CRIMINAL CONVICTION, CRIMINAL PARDON, CRIMINAL OFFENSE} & 0.25\% \\
    \hline
    [Dialect] & \texttt{DIALECT} &  2.03\% \\
    \hline
    [Disability] & \texttt{DISABILITY} & 1.80\% \\
    \hline
    [Divorce] & \texttt{DIVORCE} & 2.62\% \\
    \hline
    [Drug Test Result] & \texttt{DRUG TEST RESULT} & 2.45\% \\
    \hline
    [Employment History] & \texttt{EMPLOYMENT HISTORY} & 2.82\% \\
    \hline
    [Ethnicity and Ethnic Origin] & \texttt{ETHNICITY, NATIONALITY} & 1.85\% \\
    \hline
    [Family and Family Structure] & \texttt{FAMILY, FAMILY STRUCTURE} &  3.74\% \\
    \hline
    [Family Health History] & \texttt{FAMILY HEALTH HISTORY} &  2.16\% \\
    \hline
    [Favorite] & \texttt{FAVORITE} &  2.41\%  \\
    \hline
    [Favorite Color] & \texttt{FAVORITE COLOR} &   1.28\%  \\
    \hline
    [Favorite Food] & \texttt{FAVORITE FOOD} & 1.76\% \\
    \hline
    [Favorite Music] & \texttt{FAVORITE MUSIC} &  1.56\% \\
    \hline
    [Fetish] & \texttt{FETISH} &  1.19\% \\
    \hline
    [Gender] & \texttt{GENDER} & 1.76\% \\
    \hline
    [Hair Color] & \texttt{HAIR COLOR} &  1.55\% \\
    \hline
    [Health] & \texttt{HEALTH, HEALTH RECORD, MEDICAL HEALTH} & 1.94\% \\
    \hline
    [Health History] & \texttt{HEALTH HISTORY, INDIVIDUAL HEALTH HISTORY} &  1.78\% \\
    \hline
    [Height] & \texttt{HEIGHT} & 1.83\% \\
    \hline
    [House Owned] & \texttt{HOUSE OWNED} & 1.47\% \\
    \hline
    [Income Bracket] & \texttt{INCOME BRACKET} &  0.74\% \\
    \hline
    [Job] & \texttt{JOB} & 1.31\% \\
    \hline
    [Language] & \texttt{LANGUAGE} &  1.83\% \\
    \hline
    [Location] & \texttt{LOCATION, GEOGRAPHIC, DEMOGRAPHIC} & 0.86\% \\
    \hline
    [Marital Status] & \texttt{MARITAL STATUS} &   1.60\% \\
    \hline
    [Marriage] & \texttt{MARRIAGE} &  1.98\% \\
    \hline
    [Mental Health] & \texttt{MENTAL HEALTH}  & 1.67\% \\
    \hline
    [Name] & \texttt{NAME} & 1.46\% \\
    \hline
    [Offspring] & \texttt{OFFSPRING} & 2.01\% \\
    \hline
    [Ownership] & \texttt{OWNERSHIP, PERSONAL POSSESSION, PERSONAL DOCUMENTS} &  1.83\% \\
    \hline
    [Parent] & \texttt{PARENT} &  1.26\% \\
    \hline
    [Physical Characteristic and Trait] & \texttt{PHYSICAL CHARACTERISTIC, PHYSICAL TRAIT} &   1.51\% \\
    \hline
    [Physical Health] & \texttt{PHYSICAL HEALTH} &   1.55\% \\
    \hline
    [Piercing] & \texttt{PIERCING} &  1.33\% \\
    \hline
    [Political Affiliation] & \texttt{POLITICAL AFFILIATION, POLITICAL OPINION} & 1.35\% \\
    \hline
    [Prescription] & \texttt{PRESCRIPTION} & 1.49\% \\
    \hline
    [Privacy Preference] & \texttt{PRIVACY PREFERENCE} &   1.01\% \\
    \hline
    [Proclivitie] & \texttt{PROCLIVITIE} &  1.51\% \\
    \hline
    [Professional] & \texttt{PROFESSIONAL,CURRENT EMP., PAST EMP., WORK ENVIRONMENT} &  1.31\% \\
    \hline
    [Professional Certification] & \texttt{PROFESSIONAL CERTIFICATION} & 1.62\% \\
    \hline
    [Professional Evaluation] & \texttt{PROFESSIONAL EVALUATION, PERFORMANCE AT WORK, DISCIPLINARY ACTION} &   0.16\% \\
    \hline
    [ProfessionalInterview] & \texttt{ PROFESSIONAL INTERVIEW} &  1.11\% \\
    \hline
    [Race] & \texttt{RACE} &  1.55\% \\
    \hline
    [Reference] & \texttt{REFERENCE} & 1.67\% \\
    \hline
    [Relationship] & \texttt{RELATIONSHIP} & 1.69\% \\
    \hline
    [Religion] & \texttt{RELIGION} &  1.82\% \\
    \hline
    [Salary] & \texttt{SALARY} &  0.49\% \\
    \hline
    [School] & \texttt{SCHOOL, EDUCATION, EDUCATION EXPERIENCE, EDUCATION QUALIFICATION} &  1.74\% \\
    \hline
    [Sexual] & \texttt{SEXUAL} &   1.98\% \\
    \hline
    [Sexual History] & \texttt{SEXUAL HISTORY} &  1.82\% \\
    \hline
    [Sexual Preference] & \texttt{SEXUAL PREFERENCE} & 1.82\% \\
    \hline
    [Sibling] & \texttt{SIBLING} &  1.96\% \\
    \hline
    [Skin Tone] & \texttt{SKIN TONE} & 1.01\% \\
    \hline
    [Tattoo] & \texttt{TATTOO} & 1.60\% \\
    \hline
    [Weight] & \texttt{WEIGHT} &  1.82\% \\
    \hline
    [Work History] & \texttt{WORK HISTORY} & 1.73\% \\
    \hline
        \caption{Listed of labels in SPeDaC 3 and PDCs included \label{appendixA}} 
\end{xltabular}

\end{appendix}
\clearpage

\bibliographystyle{unsrtnat}
\bibliography{references}  %%% Uncomment this line and comment out the ``thebibliography'' section below to use the external .bib file (using bibtex) .

\begin{thebibliography}{81}
\providecommand{\natexlab}[1]{#1}
\providecommand{\url}[1]{\texttt{#1}}
\expandafter\ifx\csname urlstyle\endcsname\relax
  \providecommand{\doi}[1]{doi: #1}\else
  \providecommand{\doi}{doi: \begingroup \urlstyle{rm}\Url}\fi

\bibitem[Larson et~al.(2021)Larson, Oostdijk, and
  Zuiderveen~Borgesius]{Larson2021NotDS}
Martha Larson, Nelleke Oostdijk, and Frederik Zuiderveen~Borgesius.
\newblock \emph{Not Directly Stated, Not Explicitly Stored: Conversational
  Agents and the Privacy Threat of Implicit Information}, page 388–391.
\newblock Association for Computing Machinery, New York, NY, USA, 2021.
\newblock \doi{https://doi.org/10.1145/3450614.3463601}.

\bibitem[Adhikari and Panda(2018)]{Adhikari2018}
Kishalay Adhikari and Rajeev Panda.
\newblock Users' information privacy concerns and privacy protection behaviors
  in social networks.
\newblock \emph{Journal of Global Marketing}, pages 1--15, 01 2018.
\newblock \doi{10.1080/08911762.2017.1412552}.

\bibitem[Hendrickx et~al.(2021)Hendrickx, van Waterschoot, Khan, ten Bosch,
  Cucchiarini, and Strik]{Hendrickx2021}
Iris Hendrickx, Jelte van Waterschoot, Arif Khan, Louis ten Bosch, Catia
  Cucchiarini, and Helmer Strik.
\newblock Take back control: User privacy and transparency concerns in
  personalized conversational agents.
\newblock In \emph{IUI Workshops}, 2021.

\bibitem[IBM({\natexlab{a}})]{IBMBreachReport}
Cost of a data breach report, {\natexlab{a}}.
\newblock Available online:
  \url{https://www.ibm.com/downloads/cas/RDEQK07R}(accessed on December 05,
  2022).

\bibitem[Allahyari et~al.(2017)Allahyari, Pouriyeh, Assefi, Safaei, Trippe,
  Gutierrez, and Kochut]{Allahyari2017}
Mehdi Allahyari, Seyedamin Pouriyeh, Mehdi Assefi, Saied Safaei, Elizabeth~D.
  Trippe, Juan~B. Gutierrez, and Krys Kochut.
\newblock A brief survey of text mining: Classification, clustering and
  extraction techniques, 2017.

\bibitem[GDP({\natexlab{a}})]{GDPROnline}
Eu general data protection regulation (eu-gdpr), {\natexlab{a}}.
\newblock Available online: \url{https://gdpr.eu/} (accessed on December 05,
  2022).

\bibitem[GDP({\natexlab{b}})]{GDPROnlineRecital}
Eu general data protection regulation (eu-gdpr), {\natexlab{b}}.
\newblock Available online: \url{https://www.privacy-regulation.eu/en/r6.htm}
  (accessed on December 05, 2022).

\bibitem[GDP({\natexlab{c}})]{GDPROnlineArt4}
Eu general data protection regulation (eu-gdpr), {\natexlab{c}}.
\newblock Available online: \url{https://www.privacy-regulation.eu/en/4.htm}
  (accessed on December 05, 2022).

\bibitem[Agrawal and Srikant(2000)]{Agrawal2000}
Rakesh Agrawal and Ramakrishnan Srikant.
\newblock Privacy-preserving data mining.
\newblock In \emph{Proceedings of the 2000 ACM SIGMOD International Conference
  on Management of Data}, SIGMOD '00, page 439–450, New York, NY, USA, 2000.
  Association for Computing Machinery.
\newblock ISBN 1581132174.
\newblock \doi{10.1145/342009.335438}.
\newblock URL \url{https://doi.org/10.1145/342009.335438}.

\bibitem[Özkoç(2021)]{Esma2021}
Esma~Ergüner Özkoç.
\newblock Privacy preserving data mining.
\newblock In Ciza Thomas, editor, \emph{Data Mining}, chapter~3. IntechOpen,
  Rijeka, 2021.
\newblock \doi{10.5772/intechopen.99224}.
\newblock URL \url{https://doi.org/10.5772/intechopen.99224}.

\bibitem[Cheng et~al.(2015)Cheng, Roddick, Chu, and Lin]{Cheng2015}
Peng Cheng, John Roddick, Shu-Chuan Chu, and Chun-Wei Lin.
\newblock Privacy preservation through a greedy, distortion-based rule-hiding
  method.
\newblock \emph{Applied Intelligence}, 44, 05 2015.
\newblock \doi{10.1007/s10489-015-0671-0}.

\bibitem[Kargupta et~al.(2003)Kargupta, Datta, Wang, and
  Sivakumar]{Kargupta2003}
H.~Kargupta, S.~Datta, Q.~Wang, and Krishnamoorthy Sivakumar.
\newblock On the privacy preserving properties of random data perturbation
  techniques.
\newblock In \emph{Third IEEE International Conference on Data Mining}, pages
  99--106, 2003.
\newblock \doi{10.1109/ICDM.2003.1250908}.

\bibitem[Xiao et~al.(2021)Xiao, Xiao, Lu, Zhang, Yu, and Poor]{Xiao2021}
Yilin Xiao, Liang Xiao, Xiaozhen Lu, Hailu Zhang, Shui Yu, and H.~Vincent Poor.
\newblock Deep-reinforcement-learning-based user profile perturbation for
  privacy-aware recommendation.
\newblock \emph{IEEE Internet of Things Journal}, 8\penalty0 (6):\penalty0
  4560--4568, 2021.
\newblock \doi{10.1109/JIOT.2020.3027586}.

\bibitem[Lu et~al.(2014)Lu, Zhu, Liu, Liu, and Shao]{Lu2014}
Rongxing Lu, Hui Zhu, Ximeng Liu, Joseph~K. Liu, and Jun Shao.
\newblock Toward efficient and privacy-preserving computing in big data era.
\newblock \emph{IEEE Network}, 28\penalty0 (4):\penalty0 46--50, 2014.
\newblock \doi{10.1109/MNET.2014.6863131}.

\bibitem[Yang et~al.(2019)Yang, Zheng, Guo, Liu, and Chang]{Yang2019}
Yang Yang, Xianghan Zheng, Wenzhong Guo, Ximeng Liu, and Victor Chang.
\newblock Privacy-preserving smart iot-based healthcare big data storage and
  self-adaptive access control system.
\newblock \emph{Information Sciences}, 479:\penalty0 567--592, 2019.
\newblock ISSN 0020-0255.
\newblock \doi{https://doi.org/10.1016/j.ins.2018.02.005}.
\newblock URL
  \url{https://www.sciencedirect.com/science/article/pii/S0020025518300860}.

\bibitem[Sweeney(2002)]{Sweeney2002}
Latanya Sweeney.
\newblock K-anonymity: A model for protecting privacy.
\newblock 10\penalty0 (5):\penalty0 557–570, oct 2002.
\newblock ISSN 0218-4885.
\newblock \doi{10.1142/S0218488502001648}.
\newblock URL \url{https://doi.org/10.1142/S0218488502001648}.

\bibitem[Machanavajjhala et~al.(2006)Machanavajjhala, Gehrke, Kifer, and
  Venkitasubramaniam]{Machanavajjhala2006}
Ashwin Machanavajjhala, Johannes Gehrke, Daniel Kifer, and Muthuramakrishnan
  Venkitasubramaniam.
\newblock l-diversity: Privacy beyond k-anonymity.
\newblock volume~1, page~24, 01 2006.
\newblock \doi{10.1145/1217299.1217300}.

\bibitem[Li et~al.(2007)Li, Li, and Venkatasubramanian]{Li2007}
Ninghui Li, Tiancheng Li, and Suresh Venkatasubramanian.
\newblock t-closeness: Privacy beyond k-anonymity and l-diversity.
\newblock In \emph{2007 IEEE 23rd International Conference on Data
  Engineering}, pages 106--115, 2007.
\newblock \doi{10.1109/ICDE.2007.367856}.

\bibitem[Abadi et~al.(2016)Abadi, Chu, Goodfellow, McMahan, Mironov, Talwar,
  and Zhang]{Abadi2016}
Martin Abadi, Andy Chu, Ian Goodfellow, H.~Brendan McMahan, Ilya Mironov, Kunal
  Talwar, and Li~Zhang.
\newblock Deep learning with differential privacy.
\newblock In \emph{Proceedings of the 2016 ACM SIGSAC Conference on Computer
  and Communications Security}, 2016.
\newblock \doi{https://doi.org/10.1145/2976749.2978318}.

\bibitem[Lin et~al.(2021)Lin, Srivastava, Zhang, Djenouri, and
  Aloqaily]{Lin2021}
Jerry Chun-Wei Lin, Gautam Srivastava, Yuyu Zhang, Youcef Djenouri, and Moayad
  Aloqaily.
\newblock Privacy-preserving multiobjective sanitization model in 6g iot
  environments.
\newblock \emph{IEEE Internet of Things Journal}, 8\penalty0 (7):\penalty0
  5340--5349, 2021.
\newblock \doi{10.1109/JIOT.2020.3032896}.

\bibitem[Lin et~al.(2014)Lin, Hong, and Hsu]{Lin2014}
Chun-Wei Lin, Tzung-Pei Hong, and Hung-Chuan Hsu.
\newblock Reducing side effects of hiding sensitive itemsets in privacy
  preserving data mining.
\newblock \emph{TheScientificWorldJournal}, 2014:\penalty0 235837, 04 2014.
\newblock \doi{10.1155/2014/235837}.

\bibitem[Lin et~al.(2016)Lin, Wu, Fournier~Viger, Lin, Zhan, and
  Vozňák]{Lin2016}
Chun-Wei Lin, Tsu-Yang Wu, Philippe Fournier~Viger, Guo Lin, Justin Zhan, and
  Miroslav Vozňák.
\newblock Fast algorithms for hiding sensitive high-utility itemsets in
  privacy-preserving utility mining:.
\newblock \emph{Engineering Applications of Artificial Intelligence},
  55:\penalty0 269--284, 10 2016.
\newblock \doi{10.1016/j.engappai.2016.07.003}.

\bibitem[Lin et~al.(2018)Lin, Fournier-Viger, Wu, Gan, Djenouri, and
  Zhang]{Lin2018}
Jerry Chun-Wei Lin, Philippe Fournier-Viger, Lintai Wu, Wensheng Gan, Youcef
  Djenouri, and Ji~Zhang.
\newblock Ppsf: An open-source privacy-preserving and security mining
  framework.
\newblock In \emph{2018 IEEE International Conference on Data Mining Workshops
  (ICDMW)}, pages 1459--1463, 2018.
\newblock \doi{10.1109/ICDMW.2018.00208}.

\bibitem[Hart et~al.(2011)Hart, Manadhata, and Johnson]{Hart2011}
Michaela Ngaropaki~Teresa Hart, Pratyusa~K. Manadhata, and Rob Johnson.
\newblock Text classification for data loss prevention.
\newblock In \emph{Privacy Enhancing Technologies. PETS}, 2011.
\newblock \doi{https://doi.org/10.1007/978-3-642-22263-4_2}.

\bibitem[Goo()]{Google2022}
Google de-identify sensitive data tool.
\newblock Available online: https://cloud.google.com/dlp/docs/
  deidentify-sensitive-data\#api\_overview (accessed on December 05, 2022).

\bibitem[IBM({\natexlab{b}})]{IBM2022}
Ibm discover sensitive data tool, {\natexlab{b}}.
\newblock Available online: https://www.ibm.com/docs/en/guardium/10.6?
  topic=discover-sensitive-data (accessed on December 05, 2022).

\bibitem[Mic()]{Microsoft2022}
Microsoft pii detection tool, azure cognitive service.
\newblock Available online: https://docs.microsoft.com/
  en-us/azure/cognitive-services/language-service/personally-identifiable-information/
  (accessed on December 05, 2022).

\bibitem[Vaswani et~al.(2017)Vaswani, Shazeer, Parmar, Uszkoreit, Jones, Gomez,
  Kaiser, and Polosukhin]{Vaswani2017}
Ashish Vaswani, Noam Shazeer, Niki Parmar, Jakob Uszkoreit, Llion Jones,
  Aidan~N Gomez, \L~ukasz Kaiser, and Illia Polosukhin.
\newblock Attention is all you need.
\newblock In I.~Guyon, U.~Von Luxburg, S.~Bengio, H.~Wallach, R.~Fergus,
  S.~Vishwanathan, and R.~Garnett, editors, \emph{Advances in Neural
  Information Processing Systems}, 2017.
\newblock \doi{10.5555/3295222.3295349}.

\bibitem[Neerbek et~al.(2020)Neerbek, Eskildsen, Dolog, and
  Assent]{Neerbek2020}
Jan Neerbek, Morten Eskildsen, Peter Dolog, and Ira Assent.
\newblock A real-world data resource of complex sensitive sentences based on
  documents from the monsanto trial.
\newblock In \emph{Proceedings of the 12th Language Resources and Evaluation
  Conference}, May 2020.

\bibitem[Liu et~al.(2019)Liu, Ott, Goyal, Du, Joshi, Chen, Levy, Lewis,
  Zettlemoyer, and Stoyanov]{Liu2019}
Yinhan Liu, Myle Ott, Naman Goyal, Jingfei Du, Mandar Joshi, Danqi Chen, Omer
  Levy, Mike Lewis, Luke Zettlemoyer, and Veselin Stoyanov.
\newblock Roberta: {A} robustly optimized {BERT} pretraining approach.
\newblock \emph{CoRR}, abs/1907.11692, 2019.
\newblock \doi{http://arxiv.org/abs/1907.11692}.

\bibitem[Dias et~al.(2020)Dias, Boné, Ferreira, Ribeiro, and Maia]{Dias2020}
Mariana Dias, João Boné, João~C. Ferreira, Ricardo Ribeiro, and Rui Maia.
\newblock Named entity recognition for sensitive data discovery in portuguese.
\newblock \emph{Applied Sciences}, 10\penalty0 (7):\penalty0 2303, 2020.
\newblock \doi{https://doi.org/10.3390/app1007230}.

\bibitem[Guo et~al.(2021)Guo, Liu, Tang, and Huang]{Guo2021}
Yongyan Guo, Jiayong Liu, Wenwu Tang, and Cheng Huang.
\newblock Exsense: Extract sensitive information from unstructured data.
\newblock \emph{Comput. Secur.}, 102:\penalty0 102156, 2021.
\newblock \doi{https://doi.org/10.1016/j.cose.2020.102156}.

\bibitem[Garc{\'\i}a~Pablos et~al.(2020)Garc{\'\i}a~Pablos, Perez, and
  Cuadros]{Garcia2020}
Aitor Garc{\'\i}a~Pablos, Naiara Perez, and Montse Cuadros.
\newblock Sensitive data detection and classification in {S}panish clinical
  text: Experiments with {BERT}.
\newblock In \emph{Proceedings of the 12th Language Resources and Evaluation
  Conference}, May 2020.

\bibitem[Genetu and Tegegne(2021)]{Genetu2021}
Amare Genetu and Tesfa Tegegne.
\newblock Designing sensitive personal information detection and classification
  model for amharic text.
\newblock In \emph{2021 International Conference on Information and
  Communication Technology for Development for Africa (ICT4DA)}, 2021.
\newblock \doi{10.1109/ICT4DA53266.2021.9672227}.

\bibitem[Neerbek(2020)]{Neerbekthesis}
Jan Neerbek.
\newblock \emph{Sensitive Information Detection: Recursive Neural Networks for
  Encoding Context}.
\newblock PhD thesis, Department of Computer Science Aarhus University, 2020.

\bibitem[Caliskan et~al.(2014)Caliskan, Walsh, and Greenstadt]{Islam2014}
Aylin Caliskan, John~MacLaren Walsh, and Rachel Greenstadt.
\newblock Privacy detective: Detecting private information and collective
  privacy behavior in a large social network.
\newblock In \emph{Proceedings of the 13th Workshop on Privacy in the
  Electronic Society}, 2014.
\newblock \doi{https://doi.org/10.1145/2665943.2665958}.

\bibitem[Chow et~al.(2008)Chow, Golle, and Staddon]{Chow2008}
Richard Chow, Philippe Golle, and Jessica Staddon.
\newblock Detecting privacy leaks using corpus-based association rules.
\newblock In \emph{Proceedings of the ACM SIGKDD International Conference on
  Knowledge Discovery and Data Mining}, 08 2008.
\newblock \doi{10.1145/1401890.1401997}.

\bibitem[Geng et~al.(2011)Geng, You, Wang, and Liu]{Geng2011}
Liqiang Geng, Yonghua You, Yunli Wang, and Hongyu Liu.
\newblock Privacy measures for free text documents: Bridging the gap between
  theory and practice.
\newblock In Steven Furnell, Costas Lambrinoudakis, and G{\"u}nther Pernul,
  editors, \emph{Trust, Privacy and Security in Digital Business}, 2011.
\newblock \doi{10.1007/978-3-642-22890-2_14}.

\bibitem[McDonald et~al.(2017)McDonald, Macdonald, and Ounis]{Mcdonald2017}
Graham McDonald, Craig Macdonald, and Iadh Ounis.
\newblock Enhancing sensitivity classification with semantic features using
  word embeddings.
\newblock In Joemon~M Jose, Claudia Hauff, Ismail~Sengor Alt{\i}ngovde, Dawei
  Song, Dyaa Albakour, Stuart Watt, and John Tait, editors, \emph{Advances in
  Information Retrieval}, 2017.
\newblock \doi{https://doi.org/10.1145/3450614.3463601}.

\bibitem[LeCun et~al.(1989)LeCun, Boser, Denker, Henderson, Howard, Hubbard,
  and Jackel]{LeCun1989}
Yann LeCun, Bernhard Boser, John Denker, Donnie Henderson, R.~Howard, Wayne
  Hubbard, and Lawrence Jackel.
\newblock Handwritten digit recognition with a back-propagation network.
\newblock In D.~Touretzky, editor, \emph{Advances in Neural Information
  Processing Systems}, 1989.
\newblock \doi{10.5555/109230.109279}.

\bibitem[Xu et~al.(2019)Xu, Qi, Yu, Xu, Zhao, and Yuan]{Xu2019}
Guosheng Xu, Chunhao Qi, Hai Yu, Shengwei Xu, Chunlu Zhao, and Jing Yuan.
\newblock Detecting sensitive information of unstructured text using
  convolutional neural network.
\newblock \emph{2019 International Conference on Cyber-Enabled Distributed
  Computing and Knowledge Discovery (CyberC)}, pages 474--479, 2019.
\newblock \doi{10.1109/CyberC.2019.00087}.

\bibitem[Schuster and Paliwal(1997)]{Schuster1997}
M.~Schuster and K.K. Paliwal.
\newblock Bidirectional recurrent neural networks.
\newblock \emph{IEEE Transactions on Signal Processing}, 45\penalty0
  (11):\penalty0 2673--2681, 1997.
\newblock \doi{10.1109/78.650093}.

\bibitem[Lin et~al.(2020)Lin, Xu, Xu, Chen, and Sun]{Lin2020}
Yan Lin, Guosheng Xu, Guoai Xu, Yudong Chen, and Dawei Sun.
\newblock Sensitive information detection based on convolution neural network
  and bi-directional lstm.
\newblock In \emph{2020 IEEE 19th International Conference on Trust, Security
  and Privacy in Computing and Communications (TrustCom)}, 2020.
\newblock \doi{10.1109/TrustCom50675.2020.00223}.

\bibitem[Pas()]{PastebinOnline}
Pastebin.
\newblock Available online: \url{https://pastebin.com/} (accessed on December
  05, 2022).

\bibitem[Shi et~al.(2022)Shi, Shea, Chen, Zhang, Jia, and Yu]{Shi2022}
Weiyan Shi, Ryan Shea, Si~Chen, Chiyuan Zhang, Ruoxi Jia, and Zhou Yu.
\newblock Just fine-tune twice: Selective differential privacy for large
  language models, 2022.
\newblock URL \url{https://arxiv.org/abs/2204.07667}.

\bibitem[W3C()]{W3COnline}
W3c.
\newblock Available online: \url{https://www.w3.org/} (accessed on December 05,
  2022).

\bibitem[Pandit et~al.(2019)Pandit, Polleres, Bos, Brennan, Bruegger, Ekaputra,
  Fern{\'a}ndez, Hamed, Kiesling, Lizar, Schlehahn, Steyskal, and
  Wenning]{Pandit2019}
Harshvardhan~J. Pandit, Axel Polleres, Bert Bos, Rob Brennan, Bud Bruegger,
  Fajar~J. Ekaputra, Javier~D. Fern{\'a}ndez, Roghaiyeh~Gachpaz Hamed, Elmar
  Kiesling, Mark Lizar, Eva Schlehahn, Simon Steyskal, and Rigo Wenning.
\newblock Creating a vocabulary for data privacy.
\newblock In Herv{\'e} Panetto, Christophe Debruyne, Martin Hepp, Dave Lewis,
  Claudio~Agostino Ardagna, and Robert Meersman, editors, \emph{On the Move to
  Meaningful Internet Systems: OTM 2019 Conferences}, 2019.
\newblock \doi{https://doi.org/10.1007/978-3-030-33246-4_44}.

\bibitem[DPV({\natexlab{a}})]{DPVOnline}
Data privacy vocabulary (dpv), {\natexlab{a}}.
\newblock Available online: \url{https://w3c.github.io/dpv/dpv/} (accessed on
  December 20, 2022).

\bibitem[DPV({\natexlab{b}})]{DPV-PDOnline}
Dpv-pd: Extended personal data concepts for dpv, {\natexlab{b}}.
\newblock Available online: \url{https://w3c.github.io/dpv/dpv-pd/} (accessed
  on December 20, 2022).

\bibitem[Enr()]{EnronOnline}
Enron email dataset.
\newblock Available online: \url{https://www.cs.cmu.edu/~enron/} (accessed on
  December 05, 2022).

\bibitem[Mon()]{MonsantoOnline}
Monsanto papers.
\newblock Available online:
  \url{https://www.baumhedlundlaw.com/toxic-tort-law/monsanto-roundup-lawsuit/monsanto-papers/}
  (accessed on December 05, 2022).

\bibitem[Sánchez and Batet(2014)]{Sanchez2014a}
David Sánchez and Montserrat Batet.
\newblock C-sanitized: A privacy model for document redaction and sanitization:
  C-sanitized: A privacy model for document redaction and sanitization.
\newblock \emph{Journal of the Association for Information Science and
  Technology}, 67, 06 2014.
\newblock \doi{10.1002/asi.23363}.

\bibitem[Andreas et~al.(2021)Andreas, David, and Muiris]{Balzer2021}
Balzer Andreas, Mowatt David, and Woulfe Muiris.
\newblock Protecting personally identifiable information (pii) using tagging
  and persistence of pii, January 2021.

\bibitem[Jakubíček et~al.(2013)Jakubíček, Kilgarriff, Kovář, Rychlý, and
  Suchomel]{Jakubicek2013}
Miloš Jakubíček, Adam Kilgarriff, Vojtěch Kovář, Pavel Rychlý, and Vít
  Suchomel.
\newblock The tenten corpus family.
\newblock In \emph{7th International Corpus Linguistics Conference CL 2013},
  pages 125--127, Lancaster, 2013.

\bibitem[Klie et~al.(2018)Klie, Bugert, Boullosa, de~Castilho, and
  Gurevych]{Klie2018}
Jan-Christoph Klie, Michael Bugert, Beto Boullosa, Richard~Eckart de~Castilho,
  and Iryna Gurevych.
\newblock The inception platform: Machine-assisted and knowledge-oriented
  interactive annotation.
\newblock In \emph{COLING}, 2018.

\bibitem[Hayes and Krippendorff(2007)]{Hayes2007}
Andrew~F. Hayes and Klaus Krippendorff.
\newblock Answering the call for a standard reliability measure for coding
  data.
\newblock \emph{Communication Methods and Measures}, 1\penalty0 (1):\penalty0
  77--89, 2007.
\newblock \doi{10.1080/19312450709336664}.

\bibitem[Krippendorff(2006)]{Krippendorff2004}
Klaus Krippendorff.
\newblock {Reliability in Content Analysis: Some Common Misconceptions and
  Recommendations}.
\newblock \emph{Human Communication Research}, 30\penalty0 (3):\penalty0
  411--433, 01 2006.
\newblock ISSN 0360-3989.
\newblock \doi{10.1111/j.1468-2958.2004.tb00738.x}.
\newblock URL \url{https://doi.org/10.1111/j.1468-2958.2004.tb00738.x}.

\bibitem[Gagliardi(2018)]{Gagliardi2018}
Gloria Gagliardi.
\newblock Inter-annotator agreement in linguistica: una rassegna critica.
\newblock 12 2018.

\bibitem[Wang and Zhao(2012)]{Wang2012}
Lijun Wang and Xiqing Zhao.
\newblock Improved knn classification algorithms research in text
  categorization.
\newblock In \emph{2012 2nd International Conference on Consumer Electronics,
  Communications and Networks (CECNet)}, 2012.
\newblock \doi{10.1109/CECNet.2012.6201850}.

\bibitem[Pedregosa et~al.(2011)Pedregosa, Varoquaux, Gramfort, Michel, Thirion,
  Grisel, Blondel, Prettenhofer, Weiss, Dubourg, Vanderplas, Passos,
  Cournapeau, Brucher, Perrot, and Duchesnay]{Pedregosa2011}
Fabian Pedregosa, Ga\"{e}l Varoquaux, Alexandre Gramfort, Vincent Michel,
  Bertrand Thirion, Olivier Grisel, Mathieu Blondel, Peter Prettenhofer, Ron
  Weiss, Vincent Dubourg, Jake Vanderplas, Alexandre Passos, David Cournapeau,
  Matthieu Brucher, Matthieu Perrot, and \'{E}douard Duchesnay.
\newblock Scikit-learn: Machine learning in python.
\newblock \emph{J. Mach. Learn. Res.}, 12:\penalty0 2825–2830, nov 2011.
\newblock \doi{https://doi.org/10.48550/arXiv.1201.0490}.

\bibitem[Kearns(1990)]{Kearns1990}
M.~J. Kearns.
\newblock \emph{The Computational Complexity of Machine Learning}.
\newblock PhD thesis, USA, 1990.
\newblock UMI Order No: GAX89-26128.

\bibitem[Cortes and Vapnik(1995)]{Cortes1995}
Corinna Cortes and Vladimir Vapnik.
\newblock Support-vector networks.
\newblock 20:\penalty0 273--297, 1995.
\newblock \doi{https://doi.org/10.1007/BF00994018}.

\bibitem[Abdiansah and Wardoyo(2015)]{Abdiansah2015}
Abdiansah Abdiansah and Retantyo Wardoyo.
\newblock Article: Time complexity analysis of support vector machines (svm) in
  libsvm.
\newblock \emph{International Journal of Computer Applications}, 128\penalty0
  (3):\penalty0 28--34, October 2015.
\newblock Published by Foundation of Computer Science (FCS), NY, USA.

\bibitem[Bisong(2019)]{Bisong2019}
Ekaba Bisong.
\newblock \emph{Logistic Regression}, pages 243--250.
\newblock Apress, Berkeley, CA, 2019.
\newblock \doi{10.1007/978-1-4842-4470-8_20}.

\bibitem[He et~al.(2021)He, Liu, Gao, and Chen]{He2021}
Pengcheng He, Xiaodong Liu, Jianfeng Gao, and Weizhu Chen.
\newblock Deberta: Decoding-enhanced bert with disentangled attention.
\newblock \emph{ArXiv}, abs/2006.03654, 2021.
\newblock \doi{https://doi.org/10.48550/arXiv.2006.03654}.

\bibitem[Briskilal and Subalalitha(2022)]{Briskilal2022}
J~Briskilal and C.N. Subalalitha.
\newblock An ensemble model for classifying idioms and literal texts using bert
  and roberta.
\newblock \emph{Information Processing \& Management}, 59\penalty0
  (1):\penalty0 102756, 2022.
\newblock \doi{https://doi.org/10.1016/j.ipm.2021.102756}.

\bibitem[Qiu et~al.(2020)Qiu, Sun, Xu, Shao, Dai, and Huang]{Qiu2020}
Xipeng Qiu, Tianxiang Sun, Yige Xu, Yunfan Shao, Ning Dai, and Xuanjing Huang.
\newblock Pre-trained models for natural language processing: A survey.
\newblock \emph{Science China Technological Sciences}, 63:\penalty0 1872--1897,
  10 2020.
\newblock \doi{10.1007/s11431-020-1647-3}.

\bibitem[Bilal and Almazroi(2022)]{Bilal2022}
Muhammad Bilal and Abdulwahab Almazroi.
\newblock Effectiveness of fine-tuned bert model in classification of helpful
  and unhelpful online customer reviews.
\newblock \emph{Electronic Commerce Research}, pages 1--21, 04 2022.
\newblock \doi{10.1007/s10660-022-09560-w}.

\bibitem[Devlin et~al.(2019)Devlin, Chang, Lee, and Toutanova]{Devlin2019}
Jacob Devlin, Ming-Wei Chang, Kenton Lee, and Kristina Toutanova.
\newblock {BERT}: Pre-training of deep bidirectional transformers for language
  understanding.
\newblock In \emph{Proceedings of the 2019 Conference of the North {A}merican
  Chapter of the Association for Computational Linguistics: Human Language
  Technologies, Volume 1 (Long and Short Papers)}, June 2019.

\bibitem[RoB()]{RoBERTaHuggingface}
Roberta base model.
\newblock Available online: \url{https://huggingface.co/roberta-base}(accessed
  on December 05, 2022).

\bibitem[DeB()]{DeBERTaHuggingface}
Deberta base model.
\newblock Available online:
  \url{https://huggingface.co/microsoft/deberta-base}(accessed on December 05,
  2022).

\bibitem[Reimers and Gurevych(2019)]{Reimers2019}
Nils Reimers and Iryna Gurevych.
\newblock Sentence-{BERT}: Sentence embeddings using {S}iamese {BERT}-networks.
\newblock In \emph{Proceedings of the 2019 Conference on Empirical Methods in
  Natural Language Processing and the 9th International Joint Conference on
  Natural Language Processing (EMNLP-IJCNLP)}, pages 3982--3992, Hong Kong,
  China, November 2019. Association for Computational Linguistics.
\newblock \doi{10.18653/v1/D19-1410}.
\newblock URL \url{https://aclanthology.org/D19-1410}.

\bibitem[Feng et~al.(2020)Feng, Yang, Cer, Arivazhagan, and Wang]{Feng2020}
Fangxiaoyu Feng, Yinfei Yang, Daniel Cer, Naveen Arivazhagan, and Wei Wang.
\newblock Language-agnostic bert sentence embedding.
\newblock \emph{arXiv preprint arXiv:2007.01852}, 2020.
\newblock \doi{https://doi.org/10.48550/arXiv.2007.01852}.

\bibitem[Tripodi et~al.(2022)Tripodi, Blloshmi, and
  Levis~Sullam]{tripodi-etal-2022-evaluating}
Rocco Tripodi, Rexhina Blloshmi, and Simon Levis~Sullam.
\newblock Evaluating multilingual sentence representation models in a real case
  scenario.
\newblock In \emph{Proceedings of the Thirteenth Language Resources and
  Evaluation Conference}, pages 2928--2939, Marseille, France, June 2022.
  European Language Resources Association.
\newblock URL \url{https://aclanthology.org/2022.lrec-1.314}.

\bibitem[LaB()]{LaBmodel}
Labse model.
\newblock Available online:
  \url{https://huggingface.co/sentence\-transformers/LaBSE} (accessed on
  December 05, 2022).

\bibitem[Loshchilov and Hutter(2019)]{Loshchilov2018}
Ilya Loshchilov and Frank Hutter.
\newblock Decoupled weight decay regularization.
\newblock In \emph{International Conference on Learning Representations}, 2019.

\bibitem[M\"{u}ller et~al.(2019)M\"{u}ller, Kornblith, and Hinton]{Muller2019}
Rafael M\"{u}ller, Simon Kornblith, and Geoffrey Hinton.
\newblock \emph{When Does Label Smoothing Help?}
\newblock Number 422. Curran Associates Inc., Red Hook, NY, USA, 2019.

\bibitem[van~der Maaten and Hinton(2008)]{Maaten2008}
Laurens van~der Maaten and Geoffrey Hinton.
\newblock Visualizing data using t-sne.
\newblock \emph{Journal of Machine Learning Research}, 9\penalty0
  (86):\penalty0 2579--2605, 2008.
\newblock URL \url{http://jmlr.org/papers/v9/vandermaaten08a.html}.

\bibitem[Gambarelli and Gangemi(2022)]{Gambarelli2022}
Gaia Gambarelli and Aldo Gangemi.
\newblock Privaframe: A frame-based knowledge graph for sensitive personal
  data.
\newblock \emph{Big Data and Cognitive Computing}, 6\penalty0 (3), 2022.
\newblock \doi{10.3390/bdcc6030090}.

\bibitem[{\v{S}}uster et~al.(2017){\v{S}}uster, Tulkens, and
  Daelemans]{Suster2017}
Simon {\v{S}}uster, St{\'e}phan Tulkens, and Walter Daelemans.
\newblock A short review of ethical challenges in clinical natural language
  processing.
\newblock In \emph{Proceedings of the First {ACL} Workshop on Ethics in Natural
  Language Processing}, pages 80--87, Valencia, Spain, April 2017. Association
  for Computational Linguistics.
\newblock \doi{10.18653/v1/W17-1610}.

\bibitem[Weidinger et~al.(2021)Weidinger, Mellor, Rauh, Griffin, Uesato, Huang,
  Cheng, Glaese, Balle, Kasirzadeh, Kenton, Brown, Hawkins, Stepleton, Biles,
  Birhane, Haas, Rimell, Hendricks, Isaac, Legassick, Irving, and
  Gabriel]{Weidinger2021}
Laura Weidinger, John F.~J. Mellor, Maribeth Rauh, Conor Griffin, Jonathan
  Uesato, Po-Sen Huang, Myra Cheng, Mia Glaese, Borja Balle, Atoosa Kasirzadeh,
  Zachary Kenton, Sande~Minnich Brown, William~T. Hawkins, Tom Stepleton,
  Courtney Biles, Abeba Birhane, Julia Haas, Laura Rimell, Lisa~Anne Hendricks,
  William~S. Isaac, Sean Legassick, Geoffrey Irving, and Iason Gabriel.
\newblock Ethical and social risks of harm from language models.
\newblock \emph{ArXiv}, 2021.
\newblock \doi{abs/2112.04359}.

\end{thebibliography}





% \end{thebibliography}

\end{document}